\documentclass{article}

\usepackage{booktabs} 
\usepackage{balance} 

\usepackage[accepted]{icml2019}

\usepackage{thmtools}
\usepackage{thm-restate}
\usepackage{hyperref}
\usepackage{wrapfig}
\usepackage{mathtools}
\usepackage[normalem]{ulem}
\usepackage{url}
\usepackage{multirow}
\usepackage{xspace}
\usepackage{amsmath}
\usepackage{cleveref}
\usepackage{amsthm}
\usepackage{textcomp}
\usepackage{amssymb}
\usepackage{amsfonts}
\usepackage[final]{graphicx}
\usepackage{threeparttable}
\usepackage{enumitem}
\usepackage{parskip}
\usepackage{mathrsfs}
\usepackage{booktabs}
\usepackage{caption}
\usepackage{tcolorbox}
\usepackage{times}
\usepackage{graphicx}
\usepackage{bbm}
\usepackage{bm}
\usepackage{color}
\usepackage{colortbl}
\usepackage{bbold}

\usepackage{listings}

\usepackage{tikz}
\usepackage{tkz-graph}
\usetikzlibrary{positioning,matrix,calc,patterns}
\usepackage{caption}
\usepackage{subcaption}
\usepackage{booktabs}
\usepackage{makecell}
\usepackage{rotating}
\usepackage{pdflscape}
\usepackage{csquotes}
\usepackage{ulem}

\def\Figref#1{Figure~\ref{#1}}

\def\Secref#1{Section~\ref{#1}}

\def\eqref#1{equation~\ref{#1}}
\def\Eqref#1{Equation~\ref{#1}}

\def\1{\bm{1}}
\newcommand{\train}{\mathcal{D}}

\def\rvs{{\mathbf{s}}}

\def\rmZ{{\mathbf{Z}}}

\def\vh{{\bm{h}}}

\def\vy{{\bm{y}}}

\def\mA{{\bm{A}}}

\def\mG{{\bm{G}}}

\def\mW{{\bm{W}}}
\def\mX{{\bm{X}}}

\DeclareMathAlphabet{\mathsfit}{\encodingdefault}{\sfdefault}{m}{sl}
\SetMathAlphabet{\mathsfit}{bold}{\encodingdefault}{\sfdefault}{bx}{n}
\newcommand{\tens}[1]{\bm{\mathsfit{#1}}}
\def\tA{{\tens{A}}}

\def\tX{{\tens{X}}}

\def\gB{{\mathcal{B}}}

\def\gN{{\mathcal{N}}}

\def\sG{{\mathbb{G}}}

\def\sY{{\mathbb{Y}}}

\def\emX{{X}}

\newcommand{\etens}[1]{\mathsfit{#1}}

\def\etA{{\etens{A}}}

\def\etX{{\etens{X}}}

\newcommand{\Xv}{\mX^{(v)}}
\newcommand{\Xe}{\tX^{(e)}}
\newcommand{\eXv}{\emX^{(v)}}
\newcommand{\eXe}{\etX^{(e)}}

\newcommand{\AH}{Relational\xspace}

\newcommand{\tradn}[1]{p_{\mathrm{trad}, n}}

\newcommand{\Appendix}{Supplementary Material\xspace}
\newcommand*\dbar[1]{\overline{\overline{\lower0.2ex\hbox{$#1$}}}}

\newcommand{\harrow}[1]{\mathstrut\mkern2.5mu#1\mkern-11mu\raise1.6ex%
  \hbox{$\scriptscriptstyle\rightharpoonup$}}

\newcommand{\harrowStable}[1]{\overset{\rightharpoonup}{#1}}

\colorlet{darkgreen}{green!50!black}

\newcommand{\bW}{{\bf W}}

\newcommand{\bx}{{\bf x}}

\newcommand{\bh}{{\bf h}}

\newtheorem{theorem}{Theorem}[section]
\newtheorem*{theorem*}{Theorem}

\newtheorem{proposition}{Proposition}[section]
\newtheorem*{proposition*}{Proposition}

\newtheorem*{lemma*}{Lemma}

\newenvironment{restatProp}
{\restatable{proposition}{pisgd}}
{\endrestatable}

\newtheorem{remark}{Remark}[section]

\newtheorem{definition}{Definition}[section]
\renewenvironment{definition}{\refstepcounter{definition}\par\noindent\textit{Definition~\thedefinition:}\xspace}{\nobreak\hfill$\Diamond$\par}

\newcommand{\vadim}{d_{v}}
\newcommand{\eadim}{d_{e}}
\newcommand{\embdim}{d_{h}}

\newcommand{\indicator}[1]{\mathbb{1}_{#1}}

\newcommand{\expected}{E}

\newcommand{\graphVec}{\mathrm{vec}}
\newcommand{\skipNumVert}{M}
\newcommand{\skipLen}{R}
\newcommand{\skipGraph}{\mathcal{G}_{\text{skip}}}

\newcommand{\numWL}{L}

\newcommand{\powerGnn}{\alpha}
\newcommand{\idMatrix}{I_{|V|}}
\newcommand{\parensi}{(i)}
\newcommand{\binconcat}[2]{\left[#1 \Join #2\right]}

\newcommand{\inlinePath}[3]{{\scriptsize
		\tikz[scale=0.28]{
			\draw (0,0) circle (0cm);
			\draw[thin] (-.85,0.0) -- (-.4,0.0);
			\draw[thin] (.4,0.0) -- (.85,0.0);
			\draw[draw=black] (-1.25,0.0) circle (12pt) node {#1};
			\draw[draw=black] (0,0.0) circle (12pt) 
			node {#2};
			\draw[draw=black] (1.25,0.0) circle (12pt) node {#3};
		}
	}}

\newcommand{\inlineTriangle}[3]{{\scriptsize
                \tikz[scale=0.3]{
                        \draw (0,0) circle (0cm);
                        \draw[thin] (-.85,0.0) -- (-.4,0.0);
                        \draw[thin] (.4,0.0) -- (.85,0.0);
			\draw[thin] (-.85,0.0) to[out=90,in=90] (.85,0.0);
                        \draw[draw=black] (-1.25,0.0) circle (12pt) node {#1};
                        \draw[draw=black] (0,0.0) circle (12pt)
                        node {#2};
                        \draw[draw=black] (1.25,0.0) circle (12pt) node {#3};
                }
        }}

\newcommand{\sglist}{\mathscr{L}} 
\newcommand{\sgListKay}{\sglist_{k-1}(G_{k})}  
\newcommand{\sgListKpo}{\sglist_{k-1}(G_{k+1})}  

\def\COMPLETE{}

\begin{document}

\twocolumn[
\icmltitle{Relational Pooling for Graph Representations}



\icmlsetsymbol{equal}{*}

\begin{icmlauthorlist}
\icmlauthor{Ryan L. Murphy}{pustat}
\icmlauthor{Balasubramaniam Srinivasan}{pucs}
\icmlauthor{Vinayak Rao}{pustat}
\icmlauthor{Bruno Ribeiro}{pucs}
\end{icmlauthorlist}

\icmlaffiliation{pucs}{Department of Computer Science, Purdue University, West Lafayette, Indiana, USA}
\icmlaffiliation{pustat}{Department of Statistics, and} 

\icmlcorrespondingauthor{Ryan L. Murphy}{murph213@purdue.edu}

\icmlkeywords{Machine Learning, ICML}

\vskip 0.3in
]

\printAffiliationsAndNotice{}  

\begin{abstract}
This work generalizes graph neural networks (GNNs) beyond those based on the Weisfeiler-Lehman (WL) algorithm, graph Laplacians, and diffusions.  Our approach, denoted Relational Pooling (RP), draws from the theory of finite partial exchangeability to provide a framework with maximal representation power for graphs. RP can work with existing graph representation models and, somewhat counterintuitively, can make them even more powerful than the original WL isomorphism test. Additionally, RP allows architectures like Recurrent Neural Networks and Convolutional Neural Networks to be used in a theoretically sound approach for graph classification. We demonstrate improved performance of RP-based graph representations over state-of-the-art methods on a number of tasks.

\end{abstract}

%
%

\newcommand{\AHP}{RP\xspace}

\section{Introduction}
\label{s:intro}

Applications with relational graph data, such as molecule classification, social and biological network prediction, first order logic, and natural language understanding,  require an effective representation of graph structures and their attributes.
While representation learning for graph data has made tremendous progress in recent years, current schemes are unable to produce so-called \emph{most-powerful} representations that can provably distinguish all distinct graphs up to graph isomorphisms.
Consider for instance the broad class of Weisfeiler-Lehman (WL) based Graph Neural Networks (WL-GNNs)~\citep{Duvenaud2015, Kipf2016, gilmer17a,  Hamilton2017, 	velickovic2018graph, monti2017geometric,ying2018hierarchical,  xu2018how, morris2018weisfeiler}. These are unable to distinguish pairs of nonisomorphic graphs on which the standard WL isomorphism heuristic fails~\citep{cai1992optimal,xu2018how, morris2018weisfeiler}.
As graph neural networks (GNNs) are applied to increasingly more challenging problems, having a most-powerful framework for graph representation learning would be a key development in geometric deep learning~\citep{Bronstein2017}.

In this work we introduce {\em Relational Pooling} (RP), a novel framework 
with maximal representation power for any graph input.
In RP, we specify an idealized most-powerful 
representation for graphs and a framework for tractably approximating this ideal.
The ideal representation can distinguish pairs of nonisomorphic graphs even when the WL isomorphism test fails, which motivates a straightforward procedure using approximate RP -- we call this \emph{RP-GNN} -- for making GNNs more powerful.

A key inductive bias for graph representations is invariance to permutations of the adjacency matrix (graph isomorphisms), see 
\citet{Aldous1981, Diaconis2008, orbanz2015bayesian}.  Our work
differs 
 in its focus on learning representations of {\em finite but variable-size}  graphs.
In particular, given a finite but arbitrary-sized graph $G$ potentially endowed with vertex or edge features, RP outputs a representation $\dbar{f}(G)\!\!\in\! \mathbb{R}^{\embdim}$, $\embdim\!\!>\!\!0$ , that is invariant to graph isomorphisms.  
RP can learn representations for each vertex in a graph, though to simplify the exposition, we focus on learning one representation of the entire graph.

{\em Contributions.}
We make the following contributions:  
 (1) We introduce {\em Relational Pooling} (RP), a novel framework for graph representation that can be combined with any existing neural network architecture, including ones not generally associated with graphs such as Recurrent Neural Networks (RNNs). %
 (2) We prove that RP has maximal representation power for graphs and show that combining WL-GNNs with  RP 
can increase their representation power. 
  In our experiments, we classify graphs that cannot be distinguished by a state-of-the-art WL-GNN~\cite{xu2018how}. 
(3) We introduce approximation approaches that make RP computationally tractable. We demonstrate empirically that these  still lead to strong performance and can be used with RP-GNN 
to speed up graph classification when compared to traditional WL-GNNs.

%

\section{\AH Pooling}
\label{s:ahpooling}
\paragraph{Notation.}%
We consider graphs endowed with vertex and edge features. That is, let $G\!=\!(V,E,\Xv,\Xe)$ be a graph with vertices $V$, edges $E \subseteq V \times V$, vertex features stored in a $|V|\times d_v$ matrix $\Xv$, and edge features stored in a $|V| \times |V| \times d_e$ tensor $\Xe$. 
W.l.o.g, we let $V := \{1,\ldots,n\}$, choosing some arbitrary ordering of the vertices.
Unlike the vertex features $\Xv$, these vertex {\em labels} do not represent any meaningful information about the vertices, and learned graph representations should not depend upon the choice of ordering.  Formally, there always exists a bijection on $V$ (called a permutation or isomorphism) between orderings so we desire  \emph{permutation-invariant}, or equivalently, \emph{isomorphic-invariant} functions. 
%
%

In this work, we encode $G$ by two data structures: (1) a $|V| \times |V| \times (1 + \eadim)$ tensor that combines $G$'s adjacency matrix with its edge features and (2) a $|V|\times d_v$ matrix representing node features $\Xv$.
The tensor is defined as $\tA_{v, u, \cdot} = \binconcat{\indicator{(v,u)\in E}}{ \eXe_{v,u}}$ for $v,u \in V$ where $\binconcat{\cdot}{\cdot}$ denotes concatenation along the 3rd mode of the tensor, $\indicator{(\cdot)}$ denotes the indicator function, and $\eXe_{v,u}$ denotes the feature vector of edge $(v, u)$ by a slight abuse of notation. 
A permutation is bijection $\pi\!:\!V \to V$ from the label set $V$ to itself.
If vertices are relabeled by a permutation $\pi$, we represent the new adjacency tensor by $\tA_{\pi, \pi}$, where $(\tA_{\pi, \pi})_{\pi(i), \pi(j), k} = \etA_{i, j, k}$ $\forall i,j\!\!\in\!V$, $k\!\in\!\!\{1, \ldots, 1+\eadim \}$; the index $k$ over edge features is not permuted.  Similarly, the vertex features are represented by $\Xv_\pi$ where $(\Xv_\pi)_{\pi(i), l}\!=\!\!\eXv_{i, l}$, $\forall i \in V$ and $l\!\in\!\!\{1, \ldots, \vadim \}$. The \Appendix shows a concrete example and \citet{kearnes2016molecular} use a similar representation.

For bipartite graphs (e.g., consumers $\times$ products), $V$ is partitioned by $V^{(\text{r})}$ and $V^{(\text{c})}$ and a separate permutation function can be defined on each. Their encoding is similar to the above and 
  we define RP for the two different cases below.

\paragraph{Joint RP.}
Inspired by joint exchangeability ~\cite{Aldous1981,Diaconis2008, orbanz2015bayesian}, 
we define a {\em joint RP} permutation-invariant function 
of non-bipartite graphs, whether directed or undirected, as %
\begin{equation}\label{eq:jointinv}%
\dbar{f}(G) = \frac{1}{|V|!}\sum_{\pi \in \Pi_{|V|}} \harrow{f}(\tA_{\pi, \pi}, \Xv_\pi),%
\end{equation}%
where $\Pi_{|V|}$ is the set of all distinct permutations of $V$ and $\harrow{f}$ is an arbitrary (possibly {\em permutation-sensitive}) function of the graph with codomain $\mathbb{R}^{\embdim}$. Following~\citet{murphy2018janossy}, we use the notation $\dbar{\cdot}$ to denote permutation-invariant function.  Since~\Eqref{eq:jointinv} averages over all permutations of the labels $V$, $\dbar{f}$ is a permutation-invariant function and can theoretically represent any such function $\dbar{g}$ (consider $\harrow{f} = \dbar{g}$).   
We can compose $\dbar{f}$ with another function $\rho$ (outside the summation) to capture additional signal in the graph.  This can give a maximally expressive, albeit intractable, graph representation (Theorem~\ref{thm:maxExpressive}). We later discuss tractable  approximations for $\dbar{f}$ and neural network architectures 
for $\harrow{f}$.  

%
\paragraph{Separate RP.} 
RP for bipartite graphs is motivated by \emph{separate} exchangeability~\cite{Diaconis2008, orbanz2015bayesian} and is defined as
\begin{equation} \label{eq:sepinv}
\dbar{f}(G)\!=\!C\!\!\sum_{\pi \in \Pi_{|V^{\text{(r)}}|}}\sum_{\sigma \in \Pi_{|V^{\text{(c)}}|}  }\!\! \harrow{f}\!\big( \tA_{\pi, \sigma}, \mX_{\pi}^{(\text{r},v)}, \mX_{\sigma}^{(\text{c},v)}\big)
\end{equation}
where $C\!=\!(|V^\text{(r)}|! |V^\text{(c)}|!)^{-1}$ and $\pi$, $\sigma$ are permutations of 
$V^{(\text{r})}$, $V^{(\text{c})}$, respectively. Results that apply to joint RP apply to separate RP.

%
%

\subsection{Representation Power of RP}
Functions $\dbar{f}$ should be \emph{expressive} enough to learn distinct representations of nonisomorphic graphs or graphs with distinct features. 
We say $\dbar{f}(G)$ is \emph{most-powerful} or \emph{most-expressive} when $\dbar{f}(G)\!=\!\dbar{f}(G')$ iff $G$ and $G'$ are isomorphic and have the same vertex/edge features up to permutation. If $\dbar{f}$ is not most-powerful, a downstream function $\rho$ may struggle to predict different classes for nonisormorphic graphs.  

\begin{theorem} 
\label{thm:maxExpressive}
If node and edge attributes come from a finite set, then the representation $\dbar{f}(G)$ in \Eqref{eq:jointinv} is the most expressive representation of $G$, provided $\harrow{f}$ is sufficiently expressive (e.g., a universal approximator).
\end{theorem}
All proofs are shown in the \Appendix.    %
This result provides a key insight into \AHP; one can focus on building expressive functions $\harrow{f}$ that need not be permutation-invariant as the summation over permutations assures that permutation-invariance is satisfied.  
%
%
%
%
%
%
\subsection{Neural Network Architectures}%
\label{subsec:NnetArchitectures}
Since $\harrow{f}$ may be permutation sensitive, RP allows one to use a wide range of neural network architectures. 

\noindent
{\em RNNs, MLPs.}
A valid architecture 
is to vectorize the graph (concatenating node and edge features, as illustrated in the \Appendix) and learn $\harrow{f}$ over the resulting sequence.  
$\harrow{f}$ can be an RNN, like an LSTM~\citep{hochreiter1997long} or GRU~\citep{cho2014learning}, or a feedforward neural network (multilayer perceptron, MLP) with padding if different graphs have different sizes.  Concretely, %
\begin{equation*}
\dbar{f}(G) = 
\frac{1}{|V|!}  \sum_{\pi \in \Pi_{|V|}} \harrow{f}\big( \graphVec(\tA_{\pi, \pi}, \Xv_\pi)\big).
\end{equation*}%
{\em CNNs.} Convolutional neural networks (CNNs) can also be directly applied over the tensor $\tA_{\pi,\pi}$ and combined with the node features $\Xv_{\pi}$, as in%
\begin{equation}
\label{eq:RPCNN}
\dbar{f}(G)\!\!=\!\! \frac{1}{|V|!}\!\!\sum_{\pi \in \Pi_{|V|}}\!\!\!\!\mathrm{MLP}\bigg(\!\!\!\binconcat{\!\mathrm{CNN}(\tA_{\pi,\pi})\!}{\!\mathrm{MLP}( \Xv_\pi)\!}\!\!\!\bigg)\!,\!\!%
\end{equation}
where CNN denotes a 2D~\cite{lecun1989backpropagation,krizhevsky2012imagenet} if there are no edge features and a 3D CNN~\cite{ji20133d} if there are edge features, $\binconcat{\cdot}{\cdot}$ is a concatenation of the representations, and MLP is a multilayer perceptron.
Multi-resolution 3D convolutions~\citep{qi2016volumetric} can be used to map variable-sized graphs into the same sized representation for downstream layers.

\noindent
{\em GNNs.}
The function $\harrow{f}$ can also be a graph neural network (GNN), a broad class of models that use the graph $G$ itself to define the computation graph. These are permutation-invariant by design but we will show that their integration into RP can (1) make them more powerful and (2) speed up their computation via theoretically sound approximations.  The GNNs we consider follow a message-passing~\citep{gilmer17a} scheme defined by the recursion  
\begin{equation}\label{eq:WL}%
\bh^{(l)}_u =\phi^{(l)}\left(\bh^{(l-1)}_u, \text{JP}\big( (\bh^{(l-1)}_v)_{v \in \gN(u)} \big) \right),
\end{equation}%
where $\phi^{(l)}$ is a learnable function with distinct weights at each layer $1 \le l \leq \numWL$ of the computation graph, JP is a general (learnable) permutation-invariant function~\cite{murphy2018janossy},  $\gN(u)$ is the set of neighbors of $u \in V$, and $\bh^{(l)}_{u} \in \mathbb{R}^{d_{h}^{(l)}}$ is a vector describing the embedding of node $u$ at layer $l$.  $\bh^{(0)}_{u}$ is the feature vector of node $u$, $(\Xv)_{u, \cdot}$ or can be assigned a constant $c$ if $u$ has no features.  Under this framework, node embeddings can be used directly to predict node-level targets, or all node embeddings can be aggregated (via a learnable function) to form an embedding $\bh_{G}$ used for graph-wide tasks.  

There are several variations of~\Eqref{eq:WL} in the literature. \citet{Duvenaud2015} proposed using embeddings from all layers $l \in \{1, 2, \ldots, \numWL \}$ for graph classification. \citet{Hamilton2017} used a similar framework for node classification and link prediction tasks, using the embedding at the last layer, while \citet{Xu2018} extend~\citet{Hamilton2017} to once again use embeddings at all layers for node and link prediction tasks. Other improvements include attention~\cite{velickovic2018graph}.  This approach can be derived from spectral graph convolutions (e.g., \citep{Kipf2016}). More GNNs are discussed in Section~\ref{s:relatedwork}.
%

Recently, \citet{xu2018how, morris2018weisfeiler} showed that these architectures are at most as powerful as the Weisfeiler-Lehman (WL) algorithm for testing graph isomorphism~\citep{weisfeiler1968reduction}, which itself effectively follows a message-passing scheme. Accordingly, we will broadly refer to models defined by~\Eqref{eq:WL} as WL-GNNs. \citet{xu2018how} proposes a WL-GNN called {\em Graph Isomorphism Network} (GIN) which is as powerful as the WL test in graphs with discrete features. 


\paragraph{Can a WL-GNN be more powerful than the WL test?}%
WL-GNNs inherit a shortcoming from the WL test~\citep{ cai1992optimal, arvind2017graph,furer2017combinatorial, morris2018weisfeiler}; node representations $\bh^{(l)}_u$ do not encode whether two nodes have the {\em same neighbor} or {\em distinct neighbors with the same features}, limiting their ability to learn an expressive representation of the entire graph.
Consider a task where graphs represent molecules, where node features indicate atom type and edges denote the presence or absence of bonds. 
Here, the first WL-GNN layer cannot distinguish that two (say) carbon atoms have a bond with the same carbon atom or a bond to two distinct carbon atoms. 
Successive layers of the WL-GNN update node representations and the hope is that nodes eventually get unique representations (up to isomorphisms), and thus allow the WL-GNN to detect whether two nodes have the same neighbor based on the representations of their neighbors.  However, if there are too few WL-GNN layers 
 or complex cycles in the graph, the graph and its nodes will not be adequately represented. 

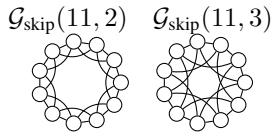
\begin{figure}[t]
\centering
\begin{tikzpicture}[auto,node distance=.8cm, main node/.style={circle,inner sep=2pt,draw,font=\sffamily\bfseries},scale=0.55] 
\node[text width=3cm] at (1, 1.5) {$\skipGraph(11, 2)$     }; 
  {some text spanning three lines with automatic line breaks};
\node[main node] (ONE) at (1.0000000 , 0.0000000) {};
\node[main node] (TWO) at (0.8412535 , 0.5406408) {};
\node[main node] (THREE) at (0.4154150 , 0.9096320) {};
\node[main node] (FOUR) at (-0.1423148 , 0.9898214) {};
\node[main node] (FIVE) at (-0.6548607 , 0.7557496) {};
\node[main node] (SIX) at (-0.9594930 , 0.2817326) {};
\node[main node] (SEVEN) at (-0.9594930 ,-0.2817326) {};
\node[main node] (EIGHT) at (-0.6548607 ,-0.7557496) {};
\node[main node] (NINE) at (-0.1423148 ,-0.9898214) {};
\node[main node] (TEN) at (0.4154150 ,-0.9096320) {};
\node[main node] (ELEVEN) at (0.8412535 ,-0.5406408) {};
%
%
\path (ONE) edge node{} (TWO);
\path (TWO) edge node{} (THREE);
\path (THREE) edge node{} (FOUR);
\path (FOUR) edge node{} (FIVE);
\path (FIVE) edge node{} (SIX);
\path (SIX) edge node{} (SEVEN);
\path (SEVEN) edge node{} (EIGHT);
\path (EIGHT) edge node{} (NINE);
\path (NINE) edge node{} (TEN);
\path (TEN) edge node{} (ELEVEN);
\path (ELEVEN) edge node{} (ONE);
%
%
\path (ONE) edge[bend left] (THREE);
\path (TWO) edge[bend left] (FOUR);
\path (THREE) edge[bend left] (FIVE);
\path (FOUR) edge[bend left] (SIX);
\path (FIVE) edge[bend left] (SEVEN);
\path (SIX) edge[bend left] (EIGHT);
\path (SEVEN) edge[bend left] (NINE);
\path (EIGHT) edge[bend left] (TEN);
\path (NINE) edge[bend left] (ELEVEN);
\path (TEN) edge[bend left] (ONE);
\path (ELEVEN) edge[bend left] (TWO);
\node[text width=3cm] at (4.5, 1.5) {$\skipGraph(11, 3)$}; 
\node[main node] (ONE) at (4.0000000 , 0.0000000) {};
\node[main node] (TWO) at (3.8412535 , 0.5406408) {};
\node[main node] (THREE) at (3.4154150 , 0.9096320) {};
\node[main node] (FOUR) at (2.85 , 0.9898214) {};
\node[main node] (FIVE) at (2.35 , 0.7557496) {};
\node[main node] (SIX) at (2.05 , 0.2817326) {};
\node[main node] (SEVEN) at (2.05 ,-0.2817326) {};
\node[main node] (EIGHT) at (2.35 ,-0.7557496) {};
\node[main node] (NINE) at (2.85 ,-0.9898214) {};
\node[main node] (TEN) at (3.4154150 ,-0.9096320) {};
\node[main node] (ELEVEN) at (3.8412535 ,-0.5406408) {};
\path (ONE) edge node{} (TWO);
\path (TWO) edge node{} (THREE);
\path (THREE) edge node{} (FOUR);
\path (FOUR) edge node{} (FIVE);
\path (FIVE) edge node{} (SIX);
\path (SIX) edge node{} (SEVEN);
\path (SEVEN) edge node{} (EIGHT);
\path (EIGHT) edge node{} (NINE);
\path (NINE) edge node{} (TEN);
\path (TEN) edge node{} (ELEVEN);
\path (ELEVEN) edge node{} (ONE);
%
%
\path (ONE) edge[bend left] (FOUR);
\path (TWO) edge[bend left] (FIVE);
\path (THREE) edge[bend left] (SIX);
\path (FOUR) edge[bend left] (SEVEN);
\path (FIVE) edge[bend left] (EIGHT);
\path (SIX) edge[bend left] (NINE);
\path (SEVEN) edge[bend left] (TEN);
\path (EIGHT) edge[bend left] (ELEVEN);
\path (NINE) edge[bend left] (ONE);
\path (TEN) edge[bend left] (TWO);
\path(ELEVEN) edge[bend left] (THREE);
\end{tikzpicture}
\caption{The WL test incorrectly deems these isomorphic.}
\label{f:circlegraphs}
\end{figure}

To better understand this challenge, consider the extreme case illustrated by the two graphs in Figure~\ref{f:circlegraphs}. 
These are cycle graphs with $\skipNumVert=11$ nodes where nodes that are $\skipLen \in \{2,3\}$ `hops' around the circle are connected by an edge. These highly symmetric graphs, which are special cases of circulant graphs~\citep{vilfredCirculant2004} are formally defined in Definition~\ref{def:circleGraph} but the key point is that the WL test, and thus WL-GNNs, cannot distinguish these two nonisomorphic graphs.  
\begin{definition}[Circulant Skip Links (CSL) graphs]
	\label{def:circleGraph}
	Let $\skipLen$ and $\skipNumVert$ be co-prime natural numbers\footnote{Two numbers are co-primes if their only common factor is 1.} such that $\skipLen < \skipNumVert-1$. $\skipGraph(\skipNumVert, \skipLen)$ denotes an undirected 4-regular graph with vertices $\{0, 1, \ldots, \skipNumVert-1 \}$ whose edges form a cycle and have skip links. That is, for the cycle,  $\{j, j+1\} \in E$ for $j \in \{0, \ldots, M-2 \}$ and $\{\skipNumVert-1, 0\}\in E$.  For the skip links, recursively define the sequence $s_{1}= 0$, $s_{i+1} = (s_{i} + \skipLen) \ \mathrm{ mod } \ \skipNumVert$ and let $\{s_{i}, s_{i+1}\} \in E$ for any $i \in \mathbb{N}$. 
\end{definition}


We will use RP to help WL-GNNs overcome this shortcoming.
Let $\harrow{f}$ be a WL-GNN that we make permutation sensitive by assigning each node an identifier that depends on $\pi$.  Permutation sensitive IDs prevent the \AHP sum from collapsing to just one term but more importantly help distinguish neighbors that otherwise appear identical. 
In particular, given any $\pi \in \Pi_{|V|}$, we append to the rows of $\Xv_{\pi}$ one-hot encodings of the row number before computing $\harrow{f}$.  We can represent this by an augmented vertex attribute matrix $\binconcat{ \Xv_\pi }{I_{|V|}}$ for every $\pi \in \Pi_{|V|}$, where $I_{|V|}$ is a $|V|\times |V|$ identity matrix and $\binconcat{B}{C}$ concatenates the columns of matrices $B$ and $C$.
RP-GNN is {then} given by %
\begin{align}\label{eq:RPGNN}
\dbar{f}(G) &= \frac{1}{|V|!}\!\sum_{\pi \in \Pi_{|V|}}\! \harrow{f}\left(\tA_{\pi, \pi}, \binconcat{\Xv_\pi} {I_{|V|}} \right) \\
			&= \frac{1}{|V|!}\!\sum_{\pi \in \Pi_{|V|}}\! \harrow{f}\left(\tA, \binconcat{\Xv} {(I_{|V|})_{\pi}} \right) \nonumber, %
\end{align}%
where the second holds since $\harrow{f}$ a GNN and thus invariant to permutations of the adjacency matrix.  
The following theorem shows that $\dbar{f}(G)$ in~\Eqref{eq:RPGNN} is strictly more \emph{expressive} than the original WL-GNN; it can distinguish all nodes and graphs that WL-GNN can in addition to graphs that the original WL-GNN cannot.  %
\begin{theorem}
	\label{p:RPGNN}
The RP-GNN in \Eqref{eq:RPGNN} is strictly more expressive than the original WL-GNN. Specifically, if $\harrow{f}$ is a GIN~\cite{xu2018how} and the graph has discrete attributes, its RP-GNN is more powerful than the WL test.
\end{theorem}%

\Eqref{eq:RPGNN} is computationally expensive but can be made tractable while retaining expressive power over standard GNNs. While all approximations discussed in Section~\ref{s:tractability} for RP in general are applicable to RP-GNN, a specific strategy is to assign permutation-sensitive node IDs in a clever way.
%
In particular, if vertex features are available, we only need to assign enough IDs to make all vertices unique and thereby reduce the number of permutations we need to evaluate.
For example, in the molecule 
$CH_2O_2$, if we create node features with one-hot IDs $(C,0,1),\!(H,0,1),\!(H,1,0),\!(O,0,1),\!(O,1,0)$, then we need only consider $1!\!\!\cdot2!\!\cdot\!2!\!\!=\!\!4$ permutations.
For unattributed graphs, we assign $i\!\!\mod m$ to node $i$; setting $m\!\!=\!\!1$ reduces to a GNN and $m\!\!=\!\!|V|$ is the most expressive. More examples are in the \Appendix.

%
%
%
%
%
%
%
%
\subsection{RP Tractability}\label{s:tractability}
\subsubsection{Tractability via canonical orientations}
\Eqref{eq:jointinv} is intractable as written and calls for approximations. 
The most direct approximation is to compose a permutation-sensitive $\harrow{f}$ with a canonical orientation function that re-orders $\tA$ such that $\text{CANONICAL}(\tA, \Xv)\!\!=\!\!\text{CANONICAL}(\tA_{\pi, \pi}, \Xv_{\pi})$, $\forall \pi \in \Pi_{|V|}$. For instance, vertices can be sorted by centrality scores with some tie-breaking scheme~\citep{montavon2012learning, niepert2016learning}.
This causes the sum over all permutations to collapse to just an evaluation of $\harrow{f} \circ \text{CANONICAL}$. 
Essentially, this introduces a fixed component into the permutation-invariant function $\dbar{f}$ with only the second stage learned from data. 
This simplifying approximation to the original problem is however only useful if $\text{CANONICAL}$ is related to the true function, and can otherwise result in poor representations \citep{murphy2018janossy}.

A more flexible approach collapses the set of all permutations into a smaller set of {\em equivalent permutations} which we denote as {\em poly-canonical orientation}. Depth-First Search (DFS) and Breadth-First Search (BFS) serve as two examples.
In a DFS, the nodes of the adjacency matrix/tensor $\tA_{\pi,\pi}$ are ordered from 1 to $|V|$ according to the order they are visited by a DFS starting at {$\pi(1)$}. 
{Thus, if $G$ is a length-three path and we consider permutation functions defined (elementwise) as $\pi(1,2,3)\!\!=\!\!(1,2,3)$, $\pi'(1,2,3)\!\!=\!\!(1,3,2)$, DFS or BFS would see respectively \inlinePath{1}{2}{3} and \inlinePath{1}{3}{2} (where vertices are numbered by permuted indices), start at $\pi(1)\!\!=\!\!1$ and result in the same `left-to-right' orientation for both permutations.} 
In disconnected graphs, the search starts at the first node of each connected component.
%
Learning orientations from data is a discrete optimization problem left for future work.
\subsubsection{Tractability via $\pi$-SGD}
A simple approach for making RP tractable is to sample {\em random} permutations during training. This offers the computational savings of a single canonical ordering but circumvents the need to learn a good canonical ordering for a given task. This approach is only approximately invariant, a tradeoff we make for the increased power of RP.

For simplicity, we analyze a supervised graph classification setting with a single sampled permutation, but this can be easily extended to sampling multiple permutations and unsupervised settings.
Further, we focus on joint invariance but the formulation is similar for separate invariance. 
Consider $N$ training data examples $\train \equiv\{(G(1),\vy(1)),\ldots,(G(N),\vy(N))\}$, where $\vy(i) \in \sY$ is the target output and graph $G(i)$ its corresponding {graph}  input. 
For a parameterized function $\harrow{f}$ with parameters $\mW$, %
\begin{equation*}\dbar{f}(G(i);\mW) =\frac{1}{|V(i)|!}\sum_{\pi \in \Pi_{|V(i)|}} \hspace{-.0in}\harrow{f}(\tA_{\pi, \pi}(i), \Xv_\pi(i);\mW),
\end{equation*} 
our (original) goal is to minimize the empirical loss %
\begin{equation}\label{eq:Loss}
\dbar{L}(\train;\mW)  = \sum_{i=1}^N \! L\left(\vy(i)\,,\,
\dbar{f}(G(i);\mW) \right),
\end{equation}%
where $L$ is a convex loss function {of $\dbar{f}(\cdot; \cdot)$} such as cross-entropy or square loss.
For each graph $G\parensi$, we sample a permutation $\rvs_i\!\sim$ $\text{Unif}(\Pi_{|V\parensi|})$ and replace the sum in \Eqref{eq:jointinv}  with the estimate%
\begin{equation}%
\hat{\dbar{f}}(G\parensi; \mW) = \harrow{f}(\tA_{\rvs_i, \rvs_i}\parensi, \Xv_{\rvs_i}\parensi;\mW).%
\label{eq:RPStochOpt}%
\end{equation}%
For separate invariance, we  
  would sample a distinct permutation for each set of vertices. The estimator in \Eqref{eq:RPStochOpt} is unbiased: $E_{\rvs_{i}}[\hat{\dbar{f}}(G_{\rvs_{i},\rvs_{i}}\parensi ; \mW)]\!\!=\!\!\dbar{f}(G\parensi; \mW)$, where $G_{\rvs_{i},\rvs_{i}}$ is shorthand for a graph that has been permuted by $\rvs_{i}$.   %
  However, this is no longer true when $\dbar{f}$ is chained with a nonlinear loss $L$:
$E_{\rvs_{i}}[L(\vy\parensi,\hat{\dbar{f}}(G_{\rvs_{i},\rvs_{i}}\parensi; \mW))]\!\!\neq\!\! 
L(\vy\parensi,E_{\rvs_{i}}[\hat{\dbar{f}}(G_{\rvs_{i},\rvs_{i}}\parensi; \mW)]) $. 
Nevertheless, as we will soon justify, we follow~\citet{murphy2018janossy} and use this estimate in our optimization.
\begin{definition}[$\pi$-SGD for RP]\label{d:piSGD}
Let $\gB_{t} =\{(G(1),\vy(1)),\ldots,(G(B),\vy(B))\}$ be a mini-batch i.i.d.\ sampled uniformly from the training data $\train$ at step $t$.  To train RP with $\pi$-SGD, we follow the stochastic gradient descent update%
\begin{equation}\label{eq:piSGD}
\mW_t = \mW_{t-1}  - \eta_{t} \rmZ_t,%
\end{equation}
where 
$\rmZ_t =\frac{1}{B} \sum_{i = 1}^{B} \nabla_{\mW} L\left(\vy(i), \hat{\dbar{f}}(G(i); \mW_{t-1}) \right) $%
is the random gradient with the random permutations $\{\rvs_i\}_{i=1}^B$, ({sampled independently} $\rvs_i \sim \text{Unif}(\Pi_{|V(i)|})$ for all graphs $G(i)$ in batch $\gB_{t}$), and the learning rate is $\eta_t \in (0,1)$ s.t.\ $\lim_{t\to \infty} \eta_t = 0$, $\sum_{t=1}^\infty \eta_t  = \infty$, and $\sum_{t=1}^\infty \eta_t^2  < \infty$.  
\end{definition}
Effectively, this is a Robbins-Monro stochastic approximation algorithm of gradient descent \citep{robbins1951stochastic,bottou2012stochastic} and optimizes the modified objective
\begin{align} \label{eq:RLoss}
  \dbar{J}&(\train;\mW) 
\!\!=\!\!\frac{1}{N} \sum_{i=1}^N  E_{\rvs_i}\left[ 
 \! L\Bigg(\vy(i) , \hat{\dbar{f}}(G_{\rvs_i,\rvs_i}(i); \mW)
 \Bigg)
\right]\nonumber\\
&=\!\frac{1}{N}\!\sum_{i=1}^N  \frac{1}{|V\parensi|!}\!\sum_{\pi \in \Pi_{|V\parensi|}}\!\!\!\!L\Bigg(\!\!\vy(i), 
\hat{\dbar{f}}(G_{\pi,\pi}(i); \mW)\!\!\Bigg).
\end{align}
Observe that the expectation over permutations is now outside the loss function {(recall $\dbar{f}(G\parensi; \mW)$ in in~\Eqref{eq:Loss} is an expectation)}.
The loss in \Eqref{eq:RLoss} is also permutation-invariant, but $\pi$-SGD yields a result sensitive to the random input permutations presented to the algorithm. 
  Further, unless the function $\harrow{f}$ itself is 
  permutation-invariant ($\dbar{f} = \harrow{f}$), the optima of 
   $\dbar{J}$ are different from those of the original objective function 
  $\dbar{L}$.
Instead, if $L$ is convex {in $\dbar{f}(\cdot ; \cdot
	)$}, $\dbar{J}$ is an upper bound to $\dbar{L}$ via Jensen's inequality, and minimizing this bound forms a tractable surrogate to the original objective in \Eqref{eq:Loss}. 

The following convergence result follows from the $\pi$-SGD formulation of~\citet{murphy2018janossy}.
\begin{restatProp}\label{p:piSGD}
	$\pi$-SGD  stochastic optimization enjoys properties of almost sure convergence to optimal $\mW$ under conditions similar to SGD (listed in Supplementary).
\end{restatProp}
\begin{remark}
\label{rmk:inference}
Given fixed point $\mW^{\star}\!$ of the $\pi$-SGD optimization and a new graph $G$ at test time, we may exactly compute $E_{\rvs}[\hat{\dbar{f}}(G_{\rvs, \rvs}; \mW^{\star})] = \dbar{f}(G; \mW^{\star})$ or estimate it with $\frac{1}{m}\sum_{j = 1}^{m} \harrow{f}(G_{\rvs_{j}, \rvs_{j}}; \mW^{\star})$, where $\rvs_{1}\!\ldots,\! \rvs_{m}\!\overset{\mathrm{i.i.d.}}{\sim}\! \mathrm{Unif}\big( \Pi_{|V|} \big)$.
\end{remark}
%
%
\subsubsection{Tractability via $k$-ary dependencies}
\label{subsubsec:karytract}
\citet{murphy2018janossy} propose $k$-ary pooling whereby the computational complexity of summing over all permutations of an input sequence is reduced by considering only permutations of subsequences of size $k$.  
Inspired by this, we propose $k$-ary \AH Pooling which operates on $k$-node induced subgraphs of $G$, which corresponds to patches of size $k \times k \times (\eadim + 1)$ of $\tA$ and $k$ rows of $\Xv$.  Formally, we define $k$-ary RP in joint RP by
\begin{equation}
\label{eq:karyAH}
\
\dbar{f}^{(k)}\!\!(G; \mW)\!=\! \frac{1}{|V|!}\!\sum_{\pi\!\in\Pi_{|V|}}\!\! \harrow{f}\!\Big(\!\tA_{\pi, \pi}[1\!\!:\!\!k, 1\!\!:\!\!k, :], \Xv_\pi[1\!\!:\!\!k,\!:]; \mW\Big),
\end{equation}
\begin{figure*}[t]
	\ifdefined\COMPLETE
\else
\documentclass[12pt]{article}
\input{util} 
\input{commands}
\input{math_commands}
\input{ryan_commands}

\begin{document}
\begin{figure}
\fi 

\begin{subfigure}[b]{.5\linewidth}
\quad 
\begin{tikzpicture}[every node/.style={anchor=north,fill=white,minimum width=1cm,minimum height=1mm,scale=.93}]
\scriptsize
\centering
\matrix (mB) [draw,matrix of math nodes]
{
|[draw,fill=red!20]|A_{(3,  3,   2)} & |[draw,fill=red!20]|A_{(3,  4,  2)} & |[draw,fill=red!20]|A_{(3,  1,  2)} & A_{(3,   2, 2)}& A_{(3,   5, 2)}\\
|[draw,fill=red!20]|A_{(4,  3,   2)} & |[draw,fill=red!20]|A_{(4,  4,  2)} & |[draw,fill=red!20]|A_{(4,  1,  2)} & A_{(4,  2, 2)} & A_{(4,  5, 2)}\\
|[draw,fill=red!20]|A_{(1,  3,   2)} & |[draw,fill=red!20]|A_{(1,  4,  2)} & |[draw,fill=red!20]|A_{(1,  1,  2)} & A_{(1,  2, 2)} & A_{(1,  5, 2)}\\
A_{(2,  3,   2)} & A_{(2, 4,  2)}& A_{(2, 1,  2)} & A_{(2, 2,  2)}& A_{(2,5, 2)}\\
A_{(5,3,   2)}& A_{(5,4,  2)} & A_{(5,1,  2)} & A_{(5,2, 2)} & A_{(5,5, 2)}\\
};
\matrix (mA) [draw,matrix of math nodes] at ($(mB.south west)+(2.5, 1.5)$)
{
|[draw,fill=red!20]|A_{(3,  3,   1)} & |[draw,fill=red!20]|A_{(3,  4,  1)} & |[draw,fill=red!20]|A_{(3,  1,  1)} & A_{(3,   2, 1)}& A_{(3,   5, 1)}\\
|[draw,fill=red!20]|A_{(4,  3,   1)} & |[draw,fill=red!20]|A_{(4,  4,  1)} & |[draw,fill=red!20]|A_{(4,  1,  1)} & A_{(4,  2, 1)} & A_{(4,  5, 1)}\\
|[draw,fill=red!20]|A_{(1,  3,   1)} & |[draw,fill=red!20]|A_{(1,  4,  1)} & |[draw,fill=red!20]|A_{(1,  1,  1)} & A_{(1,  2, 1)} & A_{(1,  5, 1)}\\
A_{(2,  3,   1)} & A_{(2, 4,  1)}& A_{(2, 1,  1)} & A_{(2, 2,  1)}& A_{(2,5, 1)}\\
A_{(5,3,   1)}& A_{(5,4,  1)} & A_{(5,1,  1)} & A_{(5,2, 1)} & A_{(5,5, 1)}\\
};

\draw[solid](mA.north east)--(mB.north east);
\draw[solid](mA.north west)--(mB.north west);
\draw[solid](mA.south east)--(mB.south east);

\end{tikzpicture}
\caption*{Adjacency tensor $\tA_{\pi^{\dagger},\pi^{\dagger}}$ where  $\pi^{\dagger}(1, 2, 3, 4, 5)= (3,4,1,2,5)$ (elementwise): the top-left $3 \times 3 \times 2$ subtensor is passed to $\harrow{f}^{(3)}$.}
\end{subfigure}
\quad \quad
\begin{subfigure}[b]{.44\linewidth}
\quad \quad\quad \quad\quad \quad
\begin{tikzpicture}[auto,node distance=3cm, thick, main node/.style={circle,draw,font=\sffamily\bfseries},scale=0.7] 
\draw (7,-1) circle (0cm); 

		   \node[pattern=dots, main node][preaction={fill=red!20}] (ONE) at (0,2) {1};
		   \node[main node] (TWO) at (0,0) {2};
		   \node[pattern=dots, main node][preaction={fill=red!20}] (THREE) at (2,2) {3};
		   \node[pattern=dots, main node][preaction={fill=red!20}] (FOUR) at (2,0) {4};
		   \node[main node] (FIVE) at (3,1) {5};
		   \path (ONE) edge node{} (TWO);
		   \path[line width=1mm] (ONE) edge node{} (THREE);
		   \path(TWO) edge node{} (THREE);
  		   \path[line width=1mm] (THREE) edge node{} (FOUR);
   		   \path (FOUR) edge node{} (FIVE);
   		   \path (THREE) edge node{} (FIVE);

\end{tikzpicture}
\caption*{An example five-node graph encoded by $\tA$. We select a 3-node induced subgraph, corresponding to the top-left of $\tA_{\pi^{\dagger}, \pi^{\dagger}}$ indicated by shaded nodes and thickened edges. }
\end{subfigure}

\ifdefined\COMPLETE

\else
\end{figure}
\end{document}
\fi
	\caption{Illustration of a $k$-ary ($k=3$) RP on a 5-node graph with one-dimensional edge attributes ($\eadim\!=\!1$) and no vertex attributes.   The graph is encoded as a $5\!\times\!5\!\times 2$ tensor $\tA$. $k$-ary RP selects the top-left $k\!\times\!k$ corner of a permuted tensor $\tA_{\pi, \pi}$. }
	\label{fig:kary}
\end{figure*}
where $\tA[\cdot, \cdot, \cdot]$ denotes access to elements in the first, second, and third modes of $\tA$; $a\!\!:\!\!b$ denotes selecting elements corresponding to indices from $a$ to $b$ inclusive; and ``$:$'' by itself denotes all elements along a mode.  Thus, we permute the adjacency tensor and select fibers along the third mode from the upper left $k \times k \times (\eadim+1)$ subtensor of $\tA$ 
as well as the vertex attributes from the first $k$ rows of $\Xv_{\pi}$.  %
  An illustration is shown in Figure~\ref{fig:kary}. The graph on the right is numbered by its `original' node indices and we assume that it has no vertex features and one-dimensional edge features.  This `original' graph would be represented by a $5 \times 5 \times 2$ tensor $\tA$ where, for all pairs of vertices, 
    the front slice holds adjacency matrix information and the back slice holds edge feature information (not shown). Given the permutation function $\pi^\dagger \in \Pi_{|V|}$ defined as $\pi^{\dagger}(1, 2, 3, 4, 5)\!=\!(3,4,1,2,5)$, the permuted $\tA_{\pi^\dagger, \pi^\dagger}$ is shown on the left. Its entries show elements from $\tA$ shuffled appropriately by $\pi^\dagger$.  For $k=3$ RP, we select the upper-left $3 \times 3$ region from $\tA_{\pi^\dagger, \pi^\dagger}$,  shaded in red, and pass this  to $\harrow{f}$. This is repeated for all permutations of the vertices.  For separate RP, the formulation is similar but we can select $k_1$ and $k_2$ nodes from $V^{\text{(r)}}$ and $V^{\text{(c)}}$, respectively.

In practice, the relevant $k$-node induced subgraphs can be selected without first permuting the entire tensor $\tA$ and matrix $\Xv$.  Instead, we enumerate all subsets of size $k$ from index set $V$ and use those to index $\tA$ and $\Xv$.

More generally, we have the following conclusion:
\begin{proposition}
\label{prop:kPerms}
The RP in \Eqref{eq:karyAH} requires summing over all $k$-node induced subgraphs of $G$, thus saving computation when $k < |V|$, reducing the number of terms in the sum from $|V|!$ to $\frac{|V|!}{(|V| - k)!}$.
\end{proposition}
Fewer computations are needed if $\!\harrow{f}\!$ is made permutation-invariant over its input $k$-node induced subgraph. We now show that the expressiveness of $k$-ary RP increases with $k$.
\begin{proposition}
\label{thm:k_implies_km}
$\dbar{f}^{(k)}$ becomes strictly more expressive as $k$ increases.
That is, for any $k\!\!\in\!\mathbb{N}$, define $\mathcal{F}_{k}$ as the set of all permutation-invariant graph functions that can be represented by RP with $k$-ary dependencies.
Then, $\mathcal{F}_{k-1} \subset \mathcal{F}_k$.    

\end{proposition}

\paragraph{Further computational savings.}%
The number of $k$-node induced subgraphs can be very large for even moderate-sized graphs. The following yield additional savings.

\noindent
{\em Ignoring some subgraphs: }%
We can encode task- and model-specific knowledge by ignoring certain $k$-sized induced subgraphs, which amounts to fixing $\harrow{f}$ to 0 for these graphs.  For example, in most applications the graph structure -- and not the node features alone -- is important so we may ignore subgraphs of $k$ isolated vertices.  Such decisions  can yield substantial computational savings in sparse graphs.

\noindent
{\em Use of $\pi$-SGD:}
We can combine the $k$-ary approximation with other strategies like $\pi$-SGD and poly-canonical orientations.
For instance, a forward pass can consist of sampling a random starting vertex and running a BFS until a $k$-node induced subgraph is selected. 
Combining $\pi$-SGD and $k$-ary RP can speed up GNNs but will not provide unbiased estimates of the loss calculated with the entire graph.
Future work could explore using the MCMC finite-sample unbiased estimator of~\citet{teixeira2018graph} with RP.

%
\section{Related Work}
\label{s:relatedwork}
Our Relational Pooling framework leverages insights from Janossy Pooling~\citep{murphy2018janossy}, which learns expressive permutation-invariant functions over \emph{sequences} by approximating an average over permutation-sensitive functions with tractability strategies. The present work raises novel applications -- like RP-GNN -- that arise when pooling over permutation-sensitive functions of~\emph{graphs}.

Graph Neural Networks (GNNs) and Graph Convolutional Networks (GCNs) form an increasingly popular class of methods~\citep{
	scarselli2009graph,
	bruna2013spectral,
	Duvenaud2015,
	niepert2016learning,
	Atwood2016,
	Kipf2016,
	gilmer17a,
	monti2017geometric, 
	defferrard2016convolutional,
	Hamilton2017,
		velickovic2018graph,
	lee2018graph,
	xu2018how}. 
 Applications include chemistry, where 
  molecules are represented as graphs 
  and we seek to predict chemical properties like toxicity~\citep{Duvenaud2015, gilmer17a, lee2018graph, wu2018moleculenet, sanchez2018inverse} and document classification 
  on a citations network~\citep{HamiltonLitReview}; and many others (cf. \citet{battaglia2018relational}). 

Recently, \citet{xu2018how} and \citet{morris2018weisfeiler} show that such GNNs are at most as powerful as the standard Weisfeiler-Lehman algorithm (also known as \emph{color refinement} or \emph{naive vertex classification}~\citep{weisfeiler1968reduction, arvind2017graph, furer2017combinatorial}) for graph isomorphism testing, and can fail to distinguish between certain classes of graphs 
\citep{cai1992optimal, arvind2017graph, furer2017combinatorial}.  In \Secref{s:experiments}, we demonstrate this phenomenon and provide empirical evidence that RP can correct some of these shortcomings. 
%
Higher-order ($k$-th order) versions of the WL test (WL[$k$]) exist and operate on tuples of size $k$ from $V$ rather than on one vertex at a time \cite{furer2017combinatorial}. 
Increasing $k$ increases the capacity of WL[$k$] to distinguish nonisomorphic graphs, which can be exploited to build more powerful GNNs~\citep{morris2018weisfeiler}. 
\citet{meng2018subgraph}, introduce a WL[$k$]-type representation to predict high-order dynamics in temporal graphs. Using GNNs based on WL[k] may be able to give better $\harrow{f}$ functions for RP but we focused on providing a representation for  
more expressive than WL[1] procedures. 
%

In another direction, WL is used to construct graph kernels~\citep{shervashidze2009efficient, shervashidze2011weisfeiler}.  CNNs have also been used with graph kernels~\cite{nikolentzos2018kernel} and some GCNs can be seen as CNNs applied to single canonical orderings~\cite{niepert2016learning,defferrard2016convolutional}.  RP provides a framework for stochastic optimization over all or poly-canonical orderings.  Another line of work derives bases for permutation-invariant functions of graphs and propose learning the coefficients of basis elements from data~\citep{maron2018invariant,hartford2018deep}.

In parallel, \citet{bloem2019probabilistic} generalized permutation-invariant functions to group-action invariant functions and discuss connections to exchangeable probability distributions~\citep{de1937prevision, Diaconis2008, Aldous1981}.
  Their theory uses a checkerboard function~\cite{orbanz2015bayesian} and the left-order canonical orientation of \citet{ghahramani2006infinite} to 
orient graphs but it will fail in some cases unless graph isomorphism can be solved in polynomial time. 
 Also, as discussed, there is no guarantee that a hand-picked canonical orientation 
will perform well on all tasks. %
On the tractability side, \citet{niepert2014tractability} shows that exchangeabilty assumptions in probabilistic graphical models provide a form of $k$-ary tractability and~\citet{cohen2016group, ravanbakhsh2017equivariance} use symmetries to reduce sample complexity and save on computation.  Another development explores the universality properties of invariance-preserving neural networks and concludes some architectures are computationally intractable~\citep{maron2019universality}. Closer to RP, \citet{montavon2012learning} discusses 
random permutations but RP provides a more comprehensive framework with theoretical analysis.

%
%
\newcommand{\gitpage}{(anonymized)}
\newcommand{\reponame}{PurdueMINDS}
\section{Experiments}
\label{s:experiments}
Our first experiment shows that RP-GNN is more expressive than WL-GNN. The second evaluates RP and its approximations on molecular data. Our code is on GitHub\footnote{\scriptsize \url{https://github.com/\reponame/RelationalPooling}}.
\subsection{Testing RP-GNN vs WL-GNN}
Here we perform experiments over the CSL graphs from \Figref{f:circlegraphs}. We demonstrate empirically that WL-GNNs are limited in their power to represent them and that RP can be used to overcome this limitation.
Our experiments compare the RP-GNN of \Eqref{eq:RPGNN} using the Graph Isomorphism Network  (GIN) architecture~\cite{xu2018how} as $\harrow{f}$ against the original GIN architecture. We choose GIN as it is arguably the most powerful WL-GNN architecture.
 
%
%

For the CSL graphs, the ``skip length'' $\skipLen$  effectively defines an isomorphism class in the sense that predicting $\skipLen$ is tantamount to classifying a graph into its isomorphism class for a fixed number of vertices $\skipNumVert$.  We are interested in predicting $\skipLen$ as an assessment of \AHP{}'s ability to exploit graph structure.  We do not claim to tackle the graph isomorphism problem as we use approximate learning ($\pi$-SGD for RP).   
\paragraph{RP-GIN.} 
GIN follows the recursion of \Eqref{eq:WL}, 
replacing \text{JP} with summation and defining $\phi^{(l)}$ as a function that sums its arguments and feeds them through an MLP:%
\begin{equation*}%
\bh^{(l)}_u =\mathrm{MLP}^{(l)}\Big( (1+\epsilon^{(l)}) \bh^{(l-1)}_u +  \sum_{v\in \gN(u)} \bh^{(l-1)}_{v}  \Big),
\end{equation*}%
for $l =1,\ldots, \numWL$, where $\{\epsilon^{(l)}\}_{l=1}^{\numWL}$ can be treated as hyper-parameters {or learned parameters} (we train $\epsilon$).  This recursion yields vertex-level representations that can be mapped to a graph-level representation by summing across $\bh^{(l)}_u$ at each given $l$, then concatenating the results, as proposed by \citet{xu2018how}. 
When applying GIN directly on our CSL graphs, we assign a constant vertex attribute to all vertices in keeping with the traditional WL algorithm, as the graph is unattributed. 
Recall that RP-GIN assigns one-hot node IDs and passes the augmented graph to GIN ($\harrow{f}$) (\Eqref{eq:RPGNN}).  We cannot assign IDs with standard GIN as doing so renders it permutation-sensitive.  Further implementation and training details are in the~\Appendix.
%
%
%
\paragraph{Classifying skip lengths.}
We create a dataset of graphs from $\big\{\skipGraph(41, \skipLen)\big\}_{\skipLen}$ where $\skipLen \in \{2, 3, 4, 5, 6, 9, 11, 12, 13, 16 \}$ and predict $\skipLen$ as a discrete response. Note $\skipNumVert\!\!=\!\!41$ is the smallest such that 10 nonisomorphic $\skipGraph(\skipNumVert, \skipLen)$ can be formed; $\exists \skipLen_{1} \ne \skipLen_{2}$ such that $\skipGraph(\skipNumVert, \skipLen_{1})$ and $\skipGraph(\skipNumVert, \skipLen_{2})$ are isomorphic.  For all 10 classes, we form 15 adjacency matrices by first constructing $A^{{(\skipLen)}}$ according to Definition~\ref{def:circleGraph} and then 14 more as $A^{{(\skipLen)}}_{\pi, \pi}$ for 14 distinct permutations $\pi$.  This gives a dataset of 150 graphs.  We evaluate GIN and RP-GIN with five-fold cross validation -- with balanced classes on both training and validation -- on this task.

\begin{table}
	\small
	\centering
	\caption{RP-GNN outperforms WL-GNN in 10-class classification task. Summary of validation-set accuracy (\%).}
	\label{tab:rpgin}
	\scalebox{0.99}{
		\tabcolsep=0.11cm
		\begin{tabular}{lrrrrrr}
			\multicolumn{1}{c}{\em model} & \multicolumn{1}{c}{mean} & \multicolumn{1}{c}{median}
			&\multicolumn{1}{c}{max} & \multicolumn{1}{c}{min} & \multicolumn{1}{c}{sd} \tabularnewline
			\hline%
			RP-GIN & 37.6 & 43.3  & 53.3 & 10.0 & 12.9 \tabularnewline
			GIN & 10.0 & 10.0 & 10.0 & 10.0 & 0.0
		\end{tabular}
	}
	\vspace{-10pt}
\end{table}
The validation-set accuracies for both models are shown in Table~\ref{tab:rpgin} and \Figref{fig:visEmbeddings} in the \Appendix.  Since GIN learns the same representation for all graphs, it predicts the same class for all graphs in the validation fold, and therefore achieves random-guessing performance of 10\% accuracy.  In comparison, RP-GIN yields substantially stronger performance on all folds, demonstrating that RP-GNNs are more powerful than their WL-GNN and serving as empirical validation of Theorem~\ref{p:RPGNN}.

%
%

\newcommand{\RPduvenaud}{RP-Duvenaud\xspace}

%
%

\subsection{Predicting Molecular Properties} 
\label{s:moleculeExperiments}
Deep learning for chemical applications learns functions on graph representations of molecules and has a rich literature~\citep{Duvenaud2015, kearnes2016molecular,  gilmer17a}. 
 This domain provides challenging tasks on which to evaluate \AHP, while in other applications, different GNN models of varying sophistication often achieve similar performance~\cite{shchur2018pitfalls, murphy2018janossy, xu2018how}.  %
We chose datasets from the MoleculeNet project~\citep{wu2018moleculenet} --  which collects chemical datasets and collates the performance of various models -- that yield classification tasks and on which graph-based methods achieved superior performance\footnote{\scriptsize \url{moleculenet.ai/latest-results}, (Dec. 2018)}.  In particular, we chose MUV~\citep{rohrer2009maximum}, HIV, and Tox21~\citep{mayr2016deeptox, huang2016tox21challenge}, which contain measurements on a molecule's biological activity, ability to inhibit HIV, and qualitative toxicity, respectively.

We processed datasets with DeepChem~\citep{RamsundarDeepChem2019} and evaluated models with ROC-AUC per the MoleculeNet project. 
Molecules are encoded as graphs with 75- and 14-dimensional node and edge features. 
 Table~\ref{tab:datasets} (in Supplementary) provides more detail.  

We use the best-performing graph model reported by MoleculeNet as $\harrow{f}$ to evaluate $k$-ary RP and to explore whether RP-GNN can make it more powerful. This is a model inspired by the GNN in~\citet{Duvenaud2015}, implemented in DeepChem by ~\citet{altae2017low}, which we refer to as the `Duvenaud et al.' model.  This model is specialized for molecules; it trains a distinct weight matrix for each possible vertex degree at each layer, which would be infeasible in other domains.  
One might ask whether RP-GNN can add any power to this state-of-the-art model, which we will explore here. We evaluated GIN~\cite{xu2018how} but it was unable to outperform `Duvenaud et al'.  
Model architectures, hyperparameters, and training procedures are detailed in the \Appendix. 
\paragraph{RP-GNN}
We compare the performance of the `Duvenaud et al.' baseline to \RPduvenaud, wherein the `Duvenaud et al.' GNN is used as $\harrow{f}$ in \Eqref{eq:RPGNN}. 
We evaluate $\harrow{f}$ on the entire graph but make \RPduvenaud tractable by training with $\pi$-SGD. At inference time, we sample 20 permutations (see Remark~\ref{rmk:inference}). Additionally, we assign just enough one-hot IDs to make atoms of the same type have unique IDs (as discussed in~\Secref{subsec:NnetArchitectures}).  To quantify variability, we train over 20 random data splits. 

The results shown in Table~\ref{tab:molaucs} suggest that \RPduvenaud is more powerful than the baseline on the HIV task and similar in performance on the others.  While we bear in mind the over-confidence in the variability estimates~\citep{bengio2004no}, this provides support of our theory.  

%
%
%
%
%
\paragraph{$k$-ary RP experiments}
Next we empirically assess the tradeoffs involved in the $k$-ary dependency models -- evaluating $\harrow{f}$ on $k$-node induced subgraphs -- discussed in \Secref{subsubsec:karytract}.  
Propositions~\ref{thm:k_implies_km} and~\ref{prop:kPerms} show that expressive power and computation decrease with $k$.
Here, $\harrow{f}$ is a `Duvenaud et al. model' that operates on induced subgraphs of size $k\!\!=\!\!10, 20, 30, 40, 50$ (the percentages of molecules with more than $k$ atoms in each dataset are shown in the \Appendix). We train using $\pi$-SGD (20 inference-time samples) and
evaluate using five random train/val/test splits.  

Results are shown in Table~\ref{tab:molaucs} and Figures~\ref{fig:toxfig},~\ref{fig:hivfig}, and~\ref{fig:muvfig} in the~\Appendix.  With the Tox21 dataset, we see a steady increase in predictive performance and computation as $k$ increases.  For instance, $k$-ary with $k=10$ is 25\% faster than the baseline with mean AUC 0.687 (0.005 sd) and with $k=20$ being 10\% faster with AUC  0.755 (0.003 sd), where (sd) indicates the standard deviation over 5 bootstrapped runs.  Results level off around $k=30$.  
{For the other datasets, neither predictive performance nor computation vary significantly with $k$. 
	Overall, the molecules are quite small and we do not expect dramatic speed-ups with smaller $k$, but this enables comparing between using the entire graph and its $k$-sized induced subgraphs.
	
	
	\begin{table}
		\small
		\centering
		\caption{Evaluation of RP-GNN and $k$-ary RP where $\harrowStable{f}$ is the `Duvenaud et al.' GNN or a neural-network. 
			We show mean (standard deviation) ROC-AUC across multiple random train/val/test splits.   
			DFS indicates Depth-First Search poly-canonical orientation.}
		\label{tab:molaucs}
		\scalebox{0.9}{
			\tabcolsep=0.11cm
			\begin{tabular}{lccc}
				\multicolumn{1}{c}{\em model} &\multicolumn{1}{c}{\textbf{HIV}} &\multicolumn{1}{c}{\textbf{MUV}}
				&\multicolumn{1}{c}{\textbf{Tox21}} \tabularnewline
				\hline
				\RPduvenaud et al.   & 0.832 (0.013) & 0.794 (0.025)  & 0.799 (0.006)\tabularnewline
				Duvenaud et al. & 0.812 (0.014) & 0.798 (0.025)  & 0.794 (0.010)\tabularnewline%
				$k\!=\!50$ Duvenaud et al.& 0.818 (0.022) & 0.768 (0.014) & 0.778 (0.007)\tabularnewline
				$k\!=\!40$ Duvenaud et al.& 0.807 (0.025) & 0.776 (0.032)  & 0.783 (0.007)\tabularnewline
				$k\!=\!30$ Duvenaud et al.& 0.829 (0.024) & 0.776 (0.030)  &  0.775 (0.011)\tabularnewline
				$k\!=\!20$ Duvenaud et al.& 0.813 (0.017) & 0.777 (0.041) & 0.755 (0.003) \tabularnewline
				$k\!=\!10$ Duvenaud et al. & 0.812 (0.035) & 0.773 (0.045) & 0.687 (0.005) \tabularnewline
				CNN-DFS &  0.542 (0.004) & 0.601 (0.042)  &  0.597 (0.006) \tabularnewline
				RNN-DFS  & 0.627 (0.007) &  0.648 (0.014) &  0.748 (0.055)\tabularnewline   
			\end{tabular}
		}
		\vspace{-10pt}
	\end{table}
	\paragraph{\AHP with CNNs and RNNs.}
	
	RP permits using neural networks for $\harrow{f}$.  We explored RNNs and CNNs and report the results in Table~\ref{tab:molaucs}.  Specific details are discussed in the \Appendix. The RNN achieves reasonable performance on Tox21 and underperforms on the other tasks.  The CNN underperforms on all tasks.  Future work is needed to determine tasks where these approaches are better suited.
\section{Conclusions}
In this work, we proposed the Relational Pooling (RP) framework for graph classification and regression. RP gives ideal most-powerful, though intractable, graph representations. We proposed several approaches to tractably approximate this ideal and
showed theoretically and empirically that RP can make WL-GNNs more expressive than the WL test. RP permits neural networks like RNNs and CNNs to be brought to such problems. Our experiments evaluate RP on a number of datasets and show how our framework can be used to improve properties of state-of-the-art methods. Future directions for theoretical study include improving our understanding of the tradeoff between representation power and computational cost of our tractability strategies.

\section*{Acknowledgments}
This work was sponsored in part by the ARO, under the U.S. Army Research Laboratory contract number W911NF-09-2-0053, the Purdue Integrative Data Science Initiative and the Purdue Research foundation, the DOD through SERC under contract number HQ0034-13-D-0004 RT \#206, and the National Science Foundation under contract numbers IIS-1816499 and DMS-1812197.

\balance
\bibliography{all_bibs}

\begin{thebibliography}{67}
\providecommand{\natexlab}[1]{#1}
\providecommand{\url}[1]{\texttt{#1}}
\expandafter\ifx\csname urlstyle\endcsname\relax
  \providecommand{\doi}[1]{doi: #1}\else
  \providecommand{\doi}{doi: \begingroup \urlstyle{rm}\Url}\fi

\bibitem[Aldous(1981)]{Aldous1981}
Aldous, D.~J.
\newblock {Representations for partially exchangeable arrays of random
  variables}.
\newblock \emph{J. Multivar. Anal.}, 11\penalty0 (4):\penalty0 581--598, 1981.
\newblock ISSN 0047259X.
\newblock \doi{10.1016/0047-259X(81)90099-3}.

\bibitem[Altae-Tran et~al.(2017)Altae-Tran, Ramsundar, Pappu, and
  Pande]{altae2017low}
Altae-Tran, H., Ramsundar, B., Pappu, A.~S., and Pande, V.
\newblock Low data drug discovery with one-shot learning.
\newblock \emph{ACS central science}, 3\penalty0 (4):\penalty0 283--293, 2017.

\bibitem[Arvind et~al.(2017)Arvind, K{\"o}bler, Rattan, and
  Verbitsky]{arvind2017graph}
Arvind, V., K{\"o}bler, J., Rattan, G., and Verbitsky, O.
\newblock Graph isomorphism, color refinement, and compactness.
\newblock \emph{computational complexity}, 26\penalty0 (3):\penalty0 627--685,
  2017.

\bibitem[Atwood \& Towsley(2016)Atwood and Towsley]{Atwood2016}
Atwood, J. and Towsley, D.
\newblock Diffusion-convolutional neural networks.
\newblock In \emph{Advances in Neural Information Processing Systems}, pp.\
  1993--2001, 2016.

\bibitem[Battaglia et~al.(2018)Battaglia, Hamrick, Bapst, Sanchez-Gonzalez,
  Zambaldi, Malinowski, Tacchetti, Raposo, Santoro, Faulkner,
  et~al.]{battaglia2018relational}
Battaglia, P.~W., Hamrick, J.~B., Bapst, V., Sanchez-Gonzalez, A., Zambaldi,
  V., Malinowski, M., Tacchetti, A., Raposo, D., Santoro, A., Faulkner, R.,
  et~al.
\newblock Relational inductive biases, deep learning, and graph networks.
\newblock \emph{arXiv preprint arXiv:1806.01261}, 2018.

\bibitem[Bengio \& Grandvalet(2004)Bengio and Grandvalet]{bengio2004no}
Bengio, Y. and Grandvalet, Y.
\newblock No unbiased estimator of the variance of k-fold cross-validation.
\newblock \emph{Journal of machine learning research}, 5\penalty0
  (Sep):\penalty0 1089--1105, 2004.

\bibitem[Bloem-Reddy \& Teh(2019)Bloem-Reddy and Teh]{bloem2019probabilistic}
Bloem-Reddy, B. and Teh, Y.~W.
\newblock Probabilistic symmetry and invariant neural networks.
\newblock \emph{arXiv preprint arXiv:1901.06082}, 2019.

\bibitem[Bottou(2012)]{bottou2012stochastic}
Bottou, L.
\newblock Stochastic gradient descent tricks.
\newblock In \emph{Neural networks: Tricks of the trade}, pp.\  421--436.
  Springer, 2012.

\bibitem[Bronstein et~al.(2017)Bronstein, Bruna, LeCun, Szlam, and
  Vandergheynst]{Bronstein2017}
Bronstein, M.~M., Bruna, J., LeCun, Y., Szlam, A., and Vandergheynst, P.
\newblock {Geometric Deep Learning: Going beyond Euclidean data}.
\newblock \emph{IEEE Signal Processing Magazine}, 34\penalty0 (4):\penalty0
  18--42, jul 2017.
\newblock ISSN 1053-5888.
\newblock \doi{10.1109/MSP.2017.2693418}.
\newblock URL \url{http://ieeexplore.ieee.org/document/7974879/}.

\bibitem[Bruna et~al.(2014)Bruna, Zaremba, Szlam, and LeCun]{bruna2013spectral}
Bruna, J., Zaremba, W., Szlam, A., and LeCun, Y.
\newblock Spectral networks and locally connected networks on graphs.
\newblock In \emph{International Conference on Learning Representations}, 2014.

\bibitem[Cai et~al.(1992)Cai, F{\"u}rer, and Immerman]{cai1992optimal}
Cai, J.-Y., F{\"u}rer, M., and Immerman, N.
\newblock An optimal lower bound on the number of variables for graph
  identification.
\newblock \emph{Combinatorica}, 12\penalty0 (4):\penalty0 389--410, 1992.

\bibitem[Cho et~al.(2014)Cho, van Merrienboer, Gulcehre, Bahdanau, Bougares,
  Schwenk, and Bengio]{cho2014learning}
Cho, K., van Merrienboer, B., Gulcehre, C., Bahdanau, D., Bougares, F.,
  Schwenk, H., and Bengio, Y.
\newblock Learning phrase representations using {RNN} encoder{--}decoder for
  statistical machine translation.
\newblock In \emph{Proceedings of the 2014 Conference on Empirical Methods in
  Natural Language Processing ({EMNLP})}, pp.\  1724--1734, Doha, Qatar,
  October 2014. Association for Computational Linguistics.
\newblock \doi{10.3115/v1/D14-1179}.
\newblock URL \url{https://www.aclweb.org/anthology/D14-1179}.

\bibitem[Cohen \& Welling(2016)Cohen and Welling]{cohen2016group}
Cohen, T. and Welling, M.
\newblock Group equivariant convolutional networks.
\newblock In \emph{International conference on machine learning}, pp.\
  2990--2999, 2016.

\bibitem[De~Finetti(1937)]{de1937prevision}
De~Finetti, B.
\newblock La pr{\'e}vision: ses lois logiques, ses sources subjectives.
\newblock In \emph{Annales de l'institut Henri Poincar{\'e}}, volume~7, pp.\
  1--68, 1937.
\newblock [Translated into Enlish: H. E. Kyburg and H.E. Smokler, eds. Studies
  in Subjective Probability. \emph{Krieger} 53-118, 1980].

\bibitem[Defferrard et~al.(2016)Defferrard, Bresson, and
  Vandergheynst]{defferrard2016convolutional}
Defferrard, M., Bresson, X., and Vandergheynst, P.
\newblock Convolutional neural networks on graphs with fast localized spectral
  filtering.
\newblock In \emph{Advances in Neural Information Processing Systems}, pp.\
  3844--3852, 2016.

\bibitem[Diaconis \& Janson(2008)Diaconis and Janson]{Diaconis2008}
Diaconis, P. and Janson, S.
\newblock {Graph limits and exchangeable random graphs}.
\newblock \emph{Rend. di Mat. e delle sue Appl. Ser. VII}, 28:\penalty0 33--61,
  2008.
\newblock ISSN 1542-7951.
\newblock \doi{10.1080/15427951.2008.10129166}.
\newblock URL \url{http://arxiv.org/abs/0712.2749}.

\bibitem[Duchi et~al.(2011)Duchi, Hazan, and Singer]{duchi2011adaptive}
Duchi, J., Hazan, E., and Singer, Y.
\newblock Adaptive subgradient methods for online learning and stochastic
  optimization.
\newblock \emph{Journal of Machine Learning Research}, 12\penalty0
  (Jul):\penalty0 2121--2159, 2011.

\bibitem[Duvenaud et~al.(2015)Duvenaud, Maclaurin, Iparraguirre, Bombarell,
  Hirzel, Aspuru-Guzik, and Adams]{Duvenaud2015}
Duvenaud, D.~K., Maclaurin, D., Iparraguirre, J., Bombarell, R., Hirzel, T.,
  Aspuru-Guzik, A., and Adams, R.~P.
\newblock Convolutional networks on graphs for learning molecular fingerprints.
\newblock In \emph{Advances in neural information processing systems}, pp.\
  2224--2232, 2015.

\bibitem[F{\"u}rer(2017)]{furer2017combinatorial}
F{\"u}rer, M.
\newblock On the combinatorial power of the {W}eisfeiler-{L}ehman algorithm.
\newblock In \emph{International Conference on Algorithms and Complexity}, pp.\
   260--271. Springer, 2017.

\bibitem[Ghahramani \& Griffiths(2006)Ghahramani and
  Griffiths]{ghahramani2006infinite}
Ghahramani, Z. and Griffiths, T.~L.
\newblock Infinite latent feature models and the {I}ndian buffet process.
\newblock In \emph{Advances in neural information processing systems}, pp.\
  475--482, 2006.

\bibitem[Gilmer et~al.(2017)Gilmer, Schoenholz, Riley, Vinyals, and
  Dahl]{gilmer17a}
Gilmer, J., Schoenholz, S.~S., Riley, P.~F., Vinyals, O., and Dahl, G.~E.
\newblock Neural message passing for quantum chemistry.
\newblock In Precup, D. and Teh, Y.~W. (eds.), \emph{Proceedings of the 34th
  International Conference on Machine Learning}, volume~70 of \emph{Proceedings
  of Machine Learning Research}, pp.\  1263--1272, International Convention
  Centre, Sydney, Australia, 06--11 Aug 2017. PMLR.
\newblock URL \url{http://proceedings.mlr.press/v70/gilmer17a.html}.

\bibitem[Hamilton et~al.(2017{\natexlab{a}})Hamilton, Ying, and
  Leskovec]{Hamilton2017}
Hamilton, W., Ying, Z., and Leskovec, J.
\newblock Inductive representation learning on large graphs.
\newblock In \emph{Advances in Neural Information Processing Systems}, pp.\
  1024--1034, 2017{\natexlab{a}}.

\bibitem[Hamilton et~al.(2017{\natexlab{b}})Hamilton, Ying, and
  Leskovec]{HamiltonLitReview}
Hamilton, W.~L., Ying, R., and Leskovec, J.
\newblock Representation learning on graphs: Methods and applications.
\newblock \emph{Bulletin of the IEEE Computer Society Technical Committee on
  Data Engineering}, 40\penalty0 (3):\penalty0 52--74, 2017{\natexlab{b}}.

\bibitem[Hartford et~al.(2018)Hartford, Graham, Leyton-Brown, and
  Ravanbakhsh]{hartford2018deep}
Hartford, J., Graham, D.~R., Leyton-Brown, K., and Ravanbakhsh, S.
\newblock Deep models of interactions across sets.
\newblock \emph{arXiv preprint arXiv:1803.02879}, 2018.

\bibitem[Hochreiter \& Schmidhuber(1997)Hochreiter and
  Schmidhuber]{hochreiter1997long}
Hochreiter, S. and Schmidhuber, J.
\newblock Long short-term memory.
\newblock \emph{Neural computation}, 9\penalty0 (8):\penalty0 1735--1780, 1997.

\bibitem[Hornik et~al.(1989)Hornik, Stinchcombe, and
  White]{hornik1989multilayer}
Hornik, K., Stinchcombe, M., and White, H.
\newblock Multilayer feedforward networks are universal approximators.
\newblock \emph{Neural networks}, 2\penalty0 (5):\penalty0 359--366, 1989.

\bibitem[Huang et~al.(2016)Huang, Xia, Nguyen, Zhao, Sakamuru, Zhao, Shahane,
  Rossoshek, and Simeonov]{huang2016tox21challenge}
Huang, R., Xia, M., Nguyen, D.-T., Zhao, T., Sakamuru, S., Zhao, J., Shahane,
  S.~A., Rossoshek, A., and Simeonov, A.
\newblock Tox21challenge to build predictive models of nuclear receptor and
  stress response pathways as mediated by exposure to environmental chemicals
  and drugs.
\newblock \emph{Frontiers in Environmental Science}, 3:\penalty0 85, 2016.

\bibitem[Ji et~al.(2013)Ji, Xu, Yang, and Yu]{ji20133d}
Ji, S., Xu, W., Yang, M., and Yu, K.
\newblock 3{D} convolutional neural networks for human action recognition.
\newblock \emph{IEEE transactions on pattern analysis and machine
  intelligence}, 35\penalty0 (1):\penalty0 221--231, 2013.

\bibitem[Kearnes et~al.(2016)Kearnes, McCloskey, Berndl, Pande, and
  Riley]{kearnes2016molecular}
Kearnes, S., McCloskey, K., Berndl, M., Pande, V., and Riley, P.
\newblock Molecular graph convolutions: moving beyond fingerprints.
\newblock \emph{Journal of computer-aided molecular design}, 30\penalty0
  (8):\penalty0 595--608, 2016.

\bibitem[Kingma \& Ba(2015)Kingma and Ba]{KingmaAdam}
Kingma, D.~P. and Ba, J.~L.
\newblock {ADAM: A Method for Stochastic Optimization}.
\newblock \emph{International Conference on Learning Representations, ICLR},
  2015.

\bibitem[Kipf \& Welling(2017)Kipf and Welling]{Kipf2016}
Kipf, T. and Welling, M.
\newblock Semi-supervised classification with graph convolutional networks.
\newblock In \emph{International Conference on Learning Representations}, 2017.

\bibitem[Krizhevsky et~al.(2012)Krizhevsky, Sutskever, and
  Hinton]{krizhevsky2012imagenet}
Krizhevsky, A., Sutskever, I., and Hinton, G.~E.
\newblock Imagenet classification with deep convolutional neural networks.
\newblock In \emph{Advances in neural information processing systems}, pp.\
  1097--1105, 2012.

\bibitem[LeCun et~al.(1989)LeCun, Boser, Denker, Henderson, Howard, Hubbard,
  and Jackel]{lecun1989backpropagation}
LeCun, Y., Boser, B., Denker, J.~S., Henderson, D., Howard, R.~E., Hubbard, W.,
  and Jackel, L.~D.
\newblock Backpropagation applied to handwritten zip code recognition.
\newblock \emph{Neural computation}, 1\penalty0 (4):\penalty0 541--551, 1989.

\bibitem[Lee et~al.(2018)Lee, Rossi, and Kong]{lee2018graph}
Lee, J.~B., Rossi, R., and Kong, X.
\newblock Graph classification using structural attention.
\newblock In \emph{Proceedings of the 24th ACM SIGKDD International Conference
  on Knowledge Discovery \& Data Mining}, pp.\  1666--1674. ACM, 2018.

\bibitem[Maron et~al.(2018)Maron, Ben-Hamu, Shamir, and
  Lipman]{maron2018invariant}
Maron, H., Ben-Hamu, H., Shamir, N., and Lipman, Y.
\newblock Invariant and equivariant graph networks.
\newblock \emph{arXiv preprint arXiv:1812.09902}, 2018.

\bibitem[Maron et~al.(2019)Maron, Fetaya, Segol, and
  Lipman]{maron2019universality}
Maron, H., Fetaya, E., Segol, N., and Lipman, Y.
\newblock On the universality of invariant networks.
\newblock \emph{arXiv preprint arXiv:1901.09342}, 2019.

\bibitem[Mayr et~al.(2016)Mayr, Klambauer, Unterthiner, and
  Hochreiter]{mayr2016deeptox}
Mayr, A., Klambauer, G., Unterthiner, T., and Hochreiter, S.
\newblock Deeptox: toxicity prediction using deep learning.
\newblock \emph{Frontiers in Environmental Science}, 3:\penalty0 80, 2016.

\bibitem[Meng et~al.(2018)Meng, Mouli, Ribeiro, and Neville]{meng2018subgraph}
Meng, C., Mouli, S.~C., Ribeiro, B., and Neville, J.
\newblock Subgraph pattern neural networks for high-order graph evolution
  prediction.
\newblock In \emph{AAAI}, 2018.

\bibitem[Montavon et~al.(2012)Montavon, Hansen, Fazli, Rupp, Biegler, Ziehe,
  Tkatchenko, Lilienfeld, and M{\"u}ller]{montavon2012learning}
Montavon, G., Hansen, K., Fazli, S., Rupp, M., Biegler, F., Ziehe, A.,
  Tkatchenko, A., Lilienfeld, A.~V., and M{\"u}ller, K.-R.
\newblock Learning invariant representations of molecules for atomization
  energy prediction.
\newblock In \emph{Advances in Neural Information Processing Systems}, pp.\
  440--448, 2012.

\bibitem[Monti et~al.(2017)Monti, Boscaini, Masci, Rodola, Svoboda, and
  Bronstein]{monti2017geometric}
Monti, F., Boscaini, D., Masci, J., Rodola, E., Svoboda, J., and Bronstein,
  M.~M.
\newblock Geometric deep learning on graphs and manifolds using mixture model
  cnns.
\newblock In \emph{Proceedings of the IEEE Conference on Computer Vision and
  Pattern Recognition}, pp.\  5115--5124, 2017.

\bibitem[Morris et~al.(2019)Morris, Ritzert, Fey, Hamilton, Lenssen, Rattan,
  and Grohe]{morris2018weisfeiler}
Morris, C., Ritzert, M., Fey, M., Hamilton, W.~L., Lenssen, J.~E., Rattan, G.,
  and Grohe, M.
\newblock {Weisfeiler and Leman Go Neural: Higher-order Graph Neural Networks}.
\newblock \emph{Proceedings of the 33rd AAAI Conference on Artificial
  Intelligence}, 2019.

\bibitem[Murphy et~al.(2019)Murphy, Srinivasan, Rao, and
  Ribeiro]{murphy2018janossy}
Murphy, R.~L., Srinivasan, B., Rao, V., and Ribeiro, B.
\newblock Janossy pooling: Learning deep permutation-invariant functions for
  variable-size inputs.
\newblock In \emph{International Conference on Learning Representations}, 2019.
\newblock URL \url{https://openreview.net/forum?id=BJluy2RcFm}.

\bibitem[Niepert \& Van~den Broeck(2014)Niepert and Van~den
  Broeck]{niepert2014tractability}
Niepert, M. and Van~den Broeck, G.
\newblock Tractability through exchangeability: A new perspective on efficient
  probabilistic inference.
\newblock In \emph{AAAI}, pp.\  2467--2475, 2014.

\bibitem[Niepert et~al.(2016)Niepert, Ahmed, and Kutzkov]{niepert2016learning}
Niepert, M., Ahmed, M., and Kutzkov, K.
\newblock Learning convolutional neural networks for graphs.
\newblock In \emph{International conference on machine learning}, pp.\
  2014--2023, 2016.

\bibitem[Nikolentzos et~al.(2018)Nikolentzos, Meladianos, Tixier, Skianis, and
  Vazirgiannis]{nikolentzos2018kernel}
Nikolentzos, G., Meladianos, P., Tixier, A. J.-P., Skianis, K., and
  Vazirgiannis, M.
\newblock Kernel graph convolutional neural networks.
\newblock In \emph{International Conference on Artificial Neural Networks},
  pp.\  22--32. Springer, 2018.

\bibitem[Orbanz \& Roy(2015)Orbanz and Roy]{orbanz2015bayesian}
Orbanz, P. and Roy, D.~M.
\newblock Bayesian models of graphs, arrays and other exchangeable random
  structures.
\newblock \emph{IEEE transactions on pattern analysis and machine
  intelligence}, 37\penalty0 (2):\penalty0 437--461, 2015.

\bibitem[Qi et~al.(2016)Qi, Su, Nie{\ss}ner, Dai, Yan, and
  Guibas]{qi2016volumetric}
Qi, C.~R., Su, H., Nie{\ss}ner, M., Dai, A., Yan, M., and Guibas, L.~J.
\newblock Volumetric and multi-view {CNNs} for object classification on 3{D}
  data.
\newblock In \emph{Proceedings of the IEEE conference on computer vision and
  pattern recognition}, pp.\  5648--5656, 2016.

\bibitem[Ramsundar et~al.(2019)Ramsundar, Eastman, Leswing, Walters, and
  Pande]{RamsundarDeepChem2019}
Ramsundar, B., Eastman, P., Leswing, K., Walters, P., and Pande, V.
\newblock \emph{Deep Learning for the Life Sciences}.
\newblock O'Reilly Media, 2019.
\newblock
  \url{https://www.amazon.com/Deep-Learning-Life-Sciences-Microscopy/dp/1492039837}.

\bibitem[Ravanbakhsh et~al.(2017)Ravanbakhsh, Schneider, and
  Poczos]{ravanbakhsh2017equivariance}
Ravanbakhsh, S., Schneider, J., and Poczos, B.
\newblock Equivariance through parameter-sharing.
\newblock In \emph{Proceedings of the 34th International Conference on Machine
  Learning-Volume 70}, pp.\  2892--2901. JMLR. org, 2017.

\bibitem[Robbins \& Monro(1951)Robbins and Monro]{robbins1951stochastic}
Robbins, H. and Monro, S.
\newblock A stochastic approximation method.
\newblock \emph{The annals of mathematical statistics}, pp.\  400--407, 1951.

\bibitem[Rohrer \& Baumann(2009)Rohrer and Baumann]{rohrer2009maximum}
Rohrer, S.~G. and Baumann, K.
\newblock Maximum unbiased validation (muv) data sets for virtual screening
  based on pubchem bioactivity data.
\newblock \emph{Journal of chemical information and modeling}, 49\penalty0
  (2):\penalty0 169--184, 2009.

\bibitem[Sanchez-Lengeling \& Aspuru-Guzik(2018)Sanchez-Lengeling and
  Aspuru-Guzik]{sanchez2018inverse}
Sanchez-Lengeling, B. and Aspuru-Guzik, A.
\newblock Inverse molecular design using machine learning: Generative models
  for matter engineering.
\newblock \emph{Science}, 361\penalty0 (6400):\penalty0 360--365, 2018.

\bibitem[Scarselli et~al.(2009)Scarselli, Gori, Tsoi, Hagenbuchner, and
  Monfardini]{scarselli2009graph}
Scarselli, F., Gori, M., Tsoi, A.~C., Hagenbuchner, M., and Monfardini, G.
\newblock The graph neural network model.
\newblock \emph{IEEE Transactions on Neural Networks}, 20\penalty0
  (1):\penalty0 61--80, 2009.

\bibitem[Shchur et~al.(2018)Shchur, Mumme, Bojchevski, and
  G{\"u}nnemann]{shchur2018pitfalls}
Shchur, O., Mumme, M., Bojchevski, A., and G{\"u}nnemann, S.
\newblock Pitfalls of graph neural network evaluation.
\newblock \emph{Relational Representation Learning Workshop (R2L 2018),
  NeurIPS}, 2018.

\bibitem[Shervashidze et~al.(2009)Shervashidze, Vishwanathan, Petri, Mehlhorn,
  and Borgwardt]{shervashidze2009efficient}
Shervashidze, N., Vishwanathan, S., Petri, T., Mehlhorn, K., and Borgwardt, K.
\newblock Efficient graphlet kernels for large graph comparison.
\newblock In \emph{Artificial Intelligence and Statistics}, pp.\  488--495,
  2009.

\bibitem[Shervashidze et~al.(2011)Shervashidze, Schweitzer, Leeuwen, Mehlhorn,
  and Borgwardt]{shervashidze2011weisfeiler}
Shervashidze, N., Schweitzer, P., Leeuwen, E. J.~v., Mehlhorn, K., and
  Borgwardt, K.~M.
\newblock {W}isfeiler-{L}ehman graph kernels.
\newblock \emph{Journal of Machine Learning Research}, 12\penalty0
  (Sep):\penalty0 2539--2561, 2011.

\bibitem[Teixeira et~al.(2018)Teixeira, Cotta, Ribeiro, and
  Meira]{teixeira2018graph}
Teixeira, C.~H., Cotta, L., Ribeiro, B., and Meira, W.
\newblock Graph pattern mining and learning through user-defined relations.
\newblock In \emph{2018 IEEE International Conference on Data Mining (ICDM)},
  pp.\  1266--1271. IEEE, 2018.

\bibitem[Velickovic et~al.(2018)Velickovic, Cucurull, Casanova, Romero, Lio,
  and Bengio]{velickovic2018graph}
Velickovic, P., Cucurull, G., Casanova, A., Romero, A., Lio, P., and Bengio, Y.
\newblock Graph attention networks.
\newblock \emph{ICLR}, 2018.

\bibitem[Vilfred(2004)]{vilfredCirculant2004}
Vilfred, V.
\newblock On circulant graphs.
\newblock In Balakrishnan, R., Sethuraman, G., and Wilson, R.~J. (eds.),
  \emph{Graph Theory and its Applications}, pp.\  34--36. Narosa Publishing
  House, 2004.

\bibitem[Weisfeiler \& Lehman(1968)Weisfeiler and
  Lehman]{weisfeiler1968reduction}
Weisfeiler, B. and Lehman, A.
\newblock A reduction of a graph to a canonical form and an algebra arising
  during this reduction.
\newblock \emph{Nauchno-Technicheskaya Informatsia}, 2\penalty0 (9):\penalty0
  12--16, 1968.

\bibitem[Wu et~al.(2018)Wu, Ramsundar, Feinberg, Gomes, Geniesse, Pappu,
  Leswing, and Pande]{wu2018moleculenet}
Wu, Z., Ramsundar, B., Feinberg, E.~N., Gomes, J., Geniesse, C., Pappu, A.~S.,
  Leswing, K., and Pande, V.
\newblock Moleculenet: a benchmark for molecular machine learning.
\newblock \emph{Chemical science}, 9\penalty0 (2):\penalty0 513--530, 2018.

\bibitem[Xu et~al.(2018)Xu, Li, Tian, Sonobe, Kawarabayashi, and
  Jegelka]{Xu2018}
Xu, K., Li, C., Tian, Y., Sonobe, T., Kawarabayashi, K.-i., and Jegelka, S.
\newblock {Representation Learning on Graphs with Jumping Knowledge Networks}.
\newblock In \emph{ICML}, 2018.
\newblock URL \url{http://arxiv.org/abs/1806.03536}.

\bibitem[Xu et~al.(2019)Xu, Hu, Leskovec, and Jegelka]{xu2018how}
Xu, K., Hu, W., Leskovec, J., and Jegelka, S.
\newblock How powerful are graph neural networks?
\newblock In \emph{International Conference on Learning Representations}, 2019.
\newblock URL \url{https://openreview.net/forum?id=ryGs6iA5Km}.

\bibitem[Ying et~al.(2018)Ying, You, Morris, Ren, Hamilton, and
  Leskovec]{ying2018hierarchical}
Ying, Z., You, J., Morris, C., Ren, X., Hamilton, W., and Leskovec, J.
\newblock Hierarchical graph representation learning with differentiable
  pooling.
\newblock In \emph{Advances in Neural Information Processing Systems}, pp.\
  4800--4810, 2018.

\bibitem[Younes(1999)]{younes1999convergence}
Younes, L.
\newblock On the convergence of markovian stochastic algorithms with rapidly
  decreasing ergodicity rates.
\newblock \emph{Stochastics: An International Journal of Probability and
  Stochastic Processes}, 65\penalty0 (3-4):\penalty0 177--228, 1999.

\bibitem[Yuille(2005)]{Yuille2004}
Yuille, A.~L.
\newblock The convergence of contrastive divergences.
\newblock In \emph{Advances in neural information processing systems}, pp.\
  1593--1600, 2005.

\bibitem[Zaheer et~al.(2017)Zaheer, Kottur, Ravanbakhsh, Poczos, Salakhutdinov,
  and Smola]{Zaheer2017}
Zaheer, M., Kottur, S., Ravanbakhsh, S., Poczos, B., Salakhutdinov, R.~R., and
  Smola, A.~J.
\newblock Deep sets.
\newblock In \emph{Advances in neural information processing systems}, pp.\
  3391--3401, 2017.

\end{thebibliography}
\bibliographystyle{icml2019}

\section*{Supplementary Material of Relational Pooling}
%
%
\renewcommand\thesection{\Alph{section}}
\setcounter{section}{0}
\section{Tensor Representation and $\graphVec$ Operation on Graphs}
\label{appendix:graphvec}
We briefly provide a concrete example of the representation of graphs and the operation $\graphVec(G)$.  Consider a graph with three vertices, one edge attribute at each edge, and two vertex attributes at each vertex.  The connectivity structure and edge attributes are represented by the $3 \times 3 \times 2$ adjacency tensor $\tA$ where $A_{(i, j, 1)}$ denotes the value of the graph's adjacency matrix and $A_{(i, j, 2)}$ denotes the value of the additional edge attribute, $i, j \in V = \{1, 2, 3 \}$.   

\quad \quad \quad \quad 
\begin{tikzpicture}[every node/.style={anchor=north ,fill=white,minimum width=1cm,minimum height=1mm}]
\small
\matrix (mB) [draw,matrix of math nodes]
{
	A_{(1, 1, 2)} & A_{(1, 2, 2)} & A_{(1, 3, 2)}\\
	A_{(2, 1, 2)} & A_{(2, 2, 2)} & A_{(2, 3, 2)} \\
	A_{(3, 1, 2)} & A_{(3, 2, 2)} & A_{(3, 3, 2)} \\
};

\matrix (mA) [draw,matrix of math nodes] at ($(mB.south west)+(0, 1)$)
{
	A_{(1, 1, 1)} & A_{(1, 2, 1)} & A_{(1 ,3, 1)}  \\
	A_{(2, 1, 1)} & A_{(2, 2, 1)} & A_{(2, 3, 1)} \\
	A_{(3, 1, 1)} & A_{(3, 2, 1)} & A_{(3, 3, 1)}\\
};
\draw[solid](mA.north east)--(mB.north east);
\draw[solid](mA.north west)--(mB.north west);
\draw[solid](mA.south east)--(mB.south east);
\end{tikzpicture}

Observe that the possibility of attributed self-loops is contemplated but this representation is applicable both to graphs that have self-loops and those that do not.  The vertex attributes are represented in a matrix %
$$ \mX^{(v)} = \begin{psmallmatrix}
X_{1, 1} & X_{1, 2} \\%
X_{2, 1} & X_{2, 2} \\%
X_{3, 1} & X_{3, 2} 
\end{psmallmatrix}. $$
A simple $\graphVec$ operation is shown below.  The modeler is free to make modifications such as applying an MLP to the vertex attributes before concatenating with the edge attributes.  Representing $G$ by $\tA$ and $\Xv$,
\begin{align*} 
	& \graphVec(G) = \big(A_{(1, 1, 1)}, A_{(1, 1, 2)}, A_{(1, 2, 1)}, A_{(1, 2, 2)}, \\%
	& A_{(1, 3, 1)}, A_{(1, 3, 2)}, X_{1, 1}, X_{1, 2}, A_{(2, 1, 1)}, A_{(2, 1, 2)},\\%
	& A_{(2, 2, 1)}, A_{(2, 2, 2)}, A_{(2, 3, 1)}, A_{(2, 3, 2)}, X_{2, 1}, X_{2, 2},\\%
	& A_{(3, 1, 1)}, A_{(3, 1, 2)}, A_{(3, 2, 1)}, A_{(3, 2, 2)}, A_{(3, 3, 1)}, \\%
	& A_{(3, 3, 2)}, X_{3, 1}, X_{3, 2} \big).
\end{align*}

Starting with the first vertex, each edge attribute (including the edge indicator) is listed, then the vertex attributes are added before doing the same with subsequent vertices.  The vectorization method for $k$-ary type models is similar, except that we apply $\graphVec$ on induced subgraphs of size $k$.
%
%

\section{More on Permutation-Invariance, WL-GNN Models, and Unique Identifiers}
\label{appendix:unique}
Here we elaborate on the addition of unique identifiers to graphs and implications for WL-GNN models.  For simplicity, we consider undirected graphs with vertex attributes but no edge attributes, allowing us to simplify our notation to an adjacency matrix $\mA$ and vertex attribute matrix $\mX$.  We also consider an oversimplified model with just one GNN layer ($\numWL=1$), the following aggregation scheme
\begin{equation*}
\bh_{u} = \bx_{u} + \sum_{v \in \mathcal{N}(u)} \bx_{v}, \quad \forall u \in V,
\end{equation*}
and the following read-out function to yield a graph representation 
\begin{equation*}
\bh_{G} = \sum_{v \in V} \bh_v.
\end{equation*}
This can be expressed as $\bh_{G} = {\bf 1}^{T}(\mA+\idMatrix)\mX$ for adjacency matrix $\mA$, vertex attribute matrix $\mX$, identity matrix $\idMatrix$, and where ${\bf 1}^{T}$ is a row vector of ones.  

For instance, we may observe the following graph with endowed vertex attributes.  The numbers indicate vertex features, not labels.\\
\begin{tikzpicture}[auto,node distance=1cm, main node/.style={circle,draw,font=\sffamily\bfseries}] 
\draw (-4, 0) circle [radius=0];
\node[main node] (ONE) at (0, 2) {\scriptsize 6};
\node[main node] (TWO) at (-.5 , 0) {\scriptsize 2};
\node[main node] (THREE) at (0.5 , 0) {\scriptsize 1};
\node[main node] (FOUR) at (0 , 1) {\scriptsize 5};
\path (ONE) edge node{} (FOUR);
\path (TWO) edge node{} (FOUR);
\path (TWO) edge node{} (THREE);
\path (FOUR) edge node{} (THREE);
\end{tikzpicture}

We can represent this with an adjacency matrix and vertex attribute matrix as $$\mA = \begin{psmallmatrix}
	0 & 0 & 0 & 1\\%
	0 & 0 & 1 & 1\\%
	0 & 1 & 0 & 1 \\
	1 & 1 & 1 & 0
\end{psmallmatrix}, %
\mX = \begin{psmallmatrix}
6 \\%
2 \\%
1  \\
5 
\end{psmallmatrix}.$$

Here, ${\bf 1}^{T}(\mA+\idMatrix)\mX = 41$.  Equivalently, we might have chosen to represent this graph as %
$$\mA_{\pi, \pi} = \begin{psmallmatrix}
0 & 0 & 1 & 0\\%
0 & 0 & 1 & 1\\%
1 & 1 & 0 & 1 \\
0 & 1 & 1 & 0
\end{psmallmatrix}, %
\mX_{\pi} = \begin{psmallmatrix}
6 \\%
2 \\%
5  \\
1 
\end{psmallmatrix}$$
Here we swapped the third and fourth column of $\mX$ and the third and fourth row and column of $\mA$.  Yet again, ${\bf 1}^{T}(\mA_{\pi, \pi}+\idMatrix)\mX_{\pi} = 41$, as desired for isomorphic-invariant functions.  We have chosen to assign scalar vertex attributes, but the invariance to permutation holds for vector vertex attributes.  

Now, we propose assigning unique one-hot IDs after constructing the adjacency matrix, which corresponds to the following representations  

$$\mA = \begin{psmallmatrix}
0 & 0 & 0 & 1\\%
0 & 0 & 1 & 1\\%
0 & 1 & 0 & 1 \\
1 & 1 & 1 & 0
\end{psmallmatrix}, %
\binconcat{\mX} {\idMatrix} = \begin{psmallmatrix}
6 & 1 & 0 & 0 & 0 \\%
2 & 0 & 1 & 0 & 0\\%
1 & 0 & 0 & 1 & 0 \\
5 & 0 & 0 & 0 & 1
\end{psmallmatrix}$$

$$\mA_{\pi, \pi} = \begin{psmallmatrix}
0 & 0 & 1 & 0\\%
0 & 0 & 1 & 1\\%
1 & 1 & 0 & 1 \\
0 & 1 & 1 & 0
\end{psmallmatrix}, %
\binconcat{ \mX_{\pi} }{\idMatrix} = \begin{psmallmatrix}
6 & 1 & 0  & 0 & 0\\%
2 & 0 & 1  & 0 & 0\\%
5 & 0 & 0  & 1 & 0\\
1 & 0 & 0  & 0 & 1
\end{psmallmatrix}$$

(recall that $\binconcat{\cdot} {\cdot}$ denotes concatenation). Note that we effectively assign identifiers after constructing $\mA$ and $\mX$ (and similarly for $\mA_{\pi, \pi}$, $\mX_{\pi}$), so that the latter four columns of $\binconcat{\mX}{\idMatrix}$ and $\binconcat{\mX_{\pi}}{\idMatrix}$ are the same.  

Now, ${\bf 1}^{T}(\mA+\idMatrix)\binconcat{\mX}{\idMatrix} = (41, 2, 3, 3, 4)$ yet ${\bf 1}^{T}(\mA_{\pi, \pi}+\idMatrix)\binconcat{\mX_{\pi}}{\idMatrix} = (41, 2, 3, 4, 3)$.  This permutation sensitivity in the presence of unique IDs holds for more general WL-GNNs and not just the one considered here.  Often $\bh_G$ is fed forward through a linear or more complex layer to obtain the final graph-level prediction and this layer is usually permutation sensitive.  Thus, we apply \AHP to GNNs with unique IDs to guarantee permutation invariance; meanwhile, the intuition for using unique IDs is to better distinguish vertices and thus create a more powerful representation for the graph.
\subsection{An alternative approach to RP-GNN models}
Next we present an equivalent but alternative representation of RP-GNN models (\Eqref{eq:RPGNN}) that may be simpler to implement in practice and provide an example.  In the previous section, we described permuting the adjacency tensor and matrix of endowed vertex attributes, leaving the matrix of identifiers unchanged.  Alternatively, with $\harrow{f}$ modeled as an isomorphic-invariant Graph Neural Network, one may leave the 
former two unchanged and instead permute the matrix of identifiers.  Thus, the alternative model becomes %
\begin{equation}\label{eq:alternativeRPGNN}
\dbar{f}(G)\! = \!\frac{1}{|V|!}\!\sum_{\pi \in \Pi_{|V|}}\! \harrow{f}\left(\tA, \binconcat{\Xv}{(\idMatrix)_{\pi}} \right),
\end{equation}
where $(\idMatrix)_{\pi}$ denotes a permutation of the rows of the identity matrix.  The more tractable version discussed previously of assigning a one-hot encoding of the id $i \!\!\mod m$ to node $i \in V$, for some $m \in \{1, 2, \ldots, |V| \}$ is still applicable. In this case, we replace $(\idMatrix)_{\pi}$ with a $|V|\times m$ matrix of $m$-bit one-hot identifiers, appropriately permuted by $\pi$.

For example, consider again the graph defined by adjacency matrix $\mA$ and vertex features $\mX$ given above.  To evaluate $\harrow{f}(\tA_{\pi, \pi}, \mX_{\pi})$ when the permutation is given by $\pi(1, 2, 3, 4) = (2, 1, 3, 4)$, we could forward%
$$A = \begin{psmallmatrix}
0 & 0 & 0 & 1\\%
0 & 0 & 1 & 1\\%
0 & 1 & 0 & 1 \\
1 & 1 & 1 & 0
\end{psmallmatrix}, %
\binconcat{\mX}{(\idMatrix)_{\pi}}= \begin{psmallmatrix}
6 & 0 & 1 & 0 & 0 \\%
2 & 1 & 0 & 0 & 0\\%
1 & 0 & 0 & 1 & 0 \\
5 & 0 & 0 & 0 & 1
\end{psmallmatrix}$$
through our model of $\harrow{f}$.  Previously the first row of $\binconcat{\mX}{\idMatrix}$ was $(6, 1, 0, 0, 0)$ whereas after permutation by $\pi$ it became $(6, 0, 1, 0, 0)$, and so on.  Both formulations discussed in this section were used in our experiments.
%
%

\section*{Proof of Theorem~\ref{thm:maxExpressive}}
\begin{proof}
Let $\Omega$ be a finite set of graphs $G~=~(V, E, \Xv, \Xe) =(\tA,\Xv)$ that includes all graph topologies for any given (arbitrarily large but finite) graph order, as well as the associated vertex and edge attributes from a finite set.  Note that isomorphic graphs $G$ and $G_{\pi, \pi}$ are considered distinct elements in $\Omega$ ($G_{\pi, \pi}$ denotes a permutation of $\tA$ and $\Xv$). If $G = (\tA, \Xv)$, let $\sG(G) = \{ (\tA_{\pi, \pi}, \Xv_{\pi}) : \pi \in \Pi_{|V|}\}$ denote the set of graphs that are isomorphic to $G$ and have the same vertex and edge attribute matrices up to permutation.
Consider a classification/prediction task where $G \in \Omega$ is assigned a target value $t(G)$ from a collection of $|\Omega|$ possible values, such that $t(G)=t(G')$ iff $G'\in \sG(G)$.
Clearly, this is the most general classification task.
Moreover, by replacing the target value $t(G)$ with a probability $p(G)$ (measure), the above task also encompasses generative tasks over $\Omega$.
All we need to show is that $\dbar{f}$ of \Eqref{eq:jointinv} is sufficiently expressive for the above task.

We now consider a permutation-sensitive function $\harrow{f}$ that assigns a distinct one-hot encoding to each distinct input graph $G = (\tA, \Xv)$. 
This $\harrow{f}$ can be approximated arbitrarily well by a sufficiently expressive neural network (operating on the vector representation of the input) as these are known to be universal approximators~\cite{hornik1989multilayer}.
Now, letting $G' \in \Omega$ be arbitrary, for all $G \in \sG(G')$, we have
\begin{align*}\dbar{f}(G) &= \frac{1}{|V|!}\sum_{\pi \in \Pi_{|V|}} \harrow{f}(\tA_{\pi,\pi},\Xv_\pi)  \\
	 &= \frac{1}{|V|!}\sum_{(\tA',\mX^{\prime(v)}) \in \sG(G)} \harrow{f}(\tA',\mX^{\prime(v)})\\
	 	 &= \frac{1}{|V|!}\sum_{(\tA',\mX^{\prime(v)}) \in \sG(G')} \harrow{f}(\tA',\mX^{\prime(v)})\\
	&= \frac{1}{|V|!}\dbar{f}(G'),
\end{align*}
thus $\dbar{f}(G')$ is the unique fingerprint of the set $\sG(G')$.
Then, all we need is a function $\rho(\cdot)$ that takes the representation $\dbar{f}(G')$ and assigns the unique target value $t(G')$, satisfying the desired condition and proving that RP has maximal representation power over $\Omega$.
\end{proof}

\section*{Proof of Theorem~\ref{p:RPGNN}}

{\em Preliminaries.}  
For an $n \times d_B$  matrix $B$ and an $n \times d_C$ matrix $C$, write $D = \binconcat{B}{C}$ to denote their concatenation to form an $n \times (d_B + d_C)$ matrix $D$.
Recall that RP-GNN adds a one-hot encoding of the {\em node id} to the node features. This {\em node id} is defined as its position in the adjacency matrix or tensor $\tA_{\pi,\pi}$ for a permutation $\pi$.  Let $I_{|V|}$ be a $|V| \times |V|$ identity matrix representing the one-hot encoding vectors of node IDs $1$ to $|V|$.
We let $\powerGnn$ denote a maximally powerful WL-GNN, that is, a deep-enough WL-GNN satisfying the conditions of Theorem 3 in \citet{xu2018how}.  That is, the multiset functions for vertex aggregation and the graph-level readout are both injective over discrete node attributes. In accordance with~\citet{xu2018how}, we focus on graphs whose features live in a countable space.  Finally, we denote multisets by $\{\{ \ldots \}\}$.

\begin{proof}
We need to show that RP-GNN is strictly more expressive than any WL-GNN.  
More specifically, we will show that RP-GNN (1) maps isomorphic graphs to the same graph embedding, (2) maps nonisomorphic graphs to distinct embeddings whenever a WL-GNN does, and (3) can map pairs of nonisomorphic graphs to distinct graph embeddings even when a most-powerful WL-GNN maps them to the same embedding. Again, in the context of the proof, when we say two graphs are `isomorphic' it is understood that they have the same topology and the same vertex/edge features up to permutation.%
We suppose that all pairs of graphs have the same number of vertices, denoted by $n$; if they have different numbers of vertices it is trivial to show that both RP-GNN and WL-GNN can represent them differently. 

(1) Assume $G_1 = (\tA_1, \Xv_1)$ and $G_2 = (\tA_2, \Xv_2)$ are isomorphic graphs with the same features (up to permutation). 
Let $\binconcat{(\Xv_j)_\pi}{I_{n}}$ 
be the new (RP-GNN) features for some permutation $\pi \in \Pi_{n}$ of graph $G_{j}$ for $j \in \{1,2 \}$.  Since $G_1$ is isomorphic to $G_2$, there exists a $\pi' \in \Pi_n$ such that %
$ \tA_1 = (\tA_{2})_{\pi',\pi'}$ and $ \binconcat{\Xv_{1}}{I_{n}}\ = \binconcat{(\Xv_{2})_{\pi'}}{I_{n}}.$
Thus%
\begin{align*}
	\dbar{f}(G_1) &= \frac{1}{n!}\sum_{\pi \in \Pi_{n}}\!\! \powerGnn\Big(\!(\tA_1)_{\pi,\pi},\!\binconcat{(\Xv_{1})_\pi}{I_{n}}\!\Big) \\
	&=\frac{1}{n!}	\sum_{\pi'' \in \Pi_{n}''}\!\! \powerGnn\Big(\!(\tA_2)_{\pi'',\pi''},\!\binconcat{(\Xv_{2})_{\pi''}}{I_{n}}\!\Big)\\
	&= \dbar{f}(G_2), 
\end{align*}
where we define $\Pi_{n}'' = \{ \pi \circ \pi' : \pi \in \Pi_{n} \}$ and observe that $\Pi_{n}'' = \Pi_{n}$.
  Thus, no pairs of isomorphic graphs will be mapped to different representations by an RP-GNN $\dbar{f}$.

(2) Assume now that $G_1 = (\tA_1, \Xv_1)$ and $G_2 = (\tA_2, \Xv_2)$  are nonisomorphic graphs with discrete attributes successfully deemed nonisomorphic by the WL test.
Theorem 3 of \citet{xu2018how} implies %
\begin{equation*}
	\powerGnn(\tA_1,\Xv_1) \neq \powerGnn(\tA_2, \Xv_2).
\end{equation*}%
Next, we can always construct an $\powerGnn'$ which has the same weights for the affine transformation over the endowed attributes as $\powerGnn$ but 
zero weights for the affine transformation over the RP-specific identifiers $I_{n}$ such that
\begin{equation*} \powerGnn'\Big(\tA_{\pi,\pi},\binconcat{(\Xv)_\pi}{I_n}\Big) 
	 = \powerGnn\Big(\tA_{\pi,\pi},\Xv_\pi\Big) 
\end{equation*}
for any $G = (\tA, \Xv)$.  Note that $\powerGnn'$ is isomorphic-invariant since $\powerGnn$ is by construction; indeed, $\powerGnn'$ ignores its permutation-sensitive part.  Thus,
\begin{align*}
	\dbar{f}(G_1) &= \frac{1}{n!}\sum_{\pi \in \Pi_n} \powerGnn'\Big((\tA_{1})_{\pi, \pi }, \binconcat{(\Xv_{1})_{\pi}}{I_{n}}\Big) \\
	&= \powerGnn\Big(\tA_1, \Xv_1\Big) \\
	&\ne \powerGnn\Big(\tA_2, \Xv_2\Big) \\
	&= \frac{1}{n!}\sum_{\pi \in \Pi_n} \powerGnn'\Big((\tA_{2})_{\pi, \pi }, \binconcat{(\Xv_{2})_{\pi}}{I_{n}}\Big) \\
	&= \dbar{f}(G_2).
\end{align*}
Therefore, RP-GNN can map graphs that WL-GNNs can distinguish to different representations, completing our proof of part (2).

(3) Finally, we construct an example to show that RP-GNN is more expressive than WL-GNN. Consider the circulant graphs with different skip links in \Figref{f:circlegraphs}. We show that these two (pairwise nonisomorphic) graphs can have different representations by RP-GNN but cannot be represented as distinct by WL-GNN. 
Let $G_1=(\tA_1,\Xv)$ denote the graph $\skipGraph(\skipNumVert = 11, \skipLen = 2)$ and $G_2=(\tA_2,\Xv)$ denote the graph $\skipGraph(\skipNumVert = 11, \skipLen = 3)$, where $\Xv = c{\bf 1}$, a vector of $c \in \mathbb{R}$.
It is not hard to show that WL-GNN cannot give different representations to $G_1$ and $G_2$, as the WL test fails in these graphs \citep{arvind2017graph, cai1992optimal,furer2017combinatorial} and the most powerful WL-GNN is just as powerful as the WL test~\cite{xu2018how}.

To show that RP-GNN is capable of giving different representations, we first show that for any given permutation $\pi$ of $G_1$, there is no permutation $\pi'$ of $G_2$ such that $\powerGnn((\tA_1)_{\pi,\pi},\binconcat{(\Xv_{1})_\pi}{I_{n}}) = \powerGnn((\tA_2)_{\pi',\pi'},\binconcat{(\Xv_{2})_{\pi'}}{I_{n}})$ (note that $\powerGnn$ is a most-powerful GNN and thus not a constant function). In this part of the proof, for simplicity and without loss of generality, consider $\pi$ a permutation such that the vertices are numbered sequentially from $1, 2, \ldots, n$ clockwise around the circle in \Figref{f:circlegraphs}.
Then, node $3$ in $G_1$ has neighbors $N_3 = \{1,2,4,5\}$ and node $4$ has neighbors $N_4 = \{2,3,5,6\}$, with intersection $N_3 \cap N_4 = \{2,5\}$.  However, in $G_2$, no two nodes share two neighbors.  Therefore, the multisets of all neighborhood attribute sequences for both permuted graphs -- denoted in $(G_{l})_{\pi, \pi}$ as $\big\{\big\{ (\bh^{(0)}_{l, u})_{u \in N_{v}} \big\}\big\}_{v \in V_{l}}$,  for $l \in \{1,2 \}$, where the $\bh^{(0)}$ terms include the rows of $I_n$ -- will be distinct.  Thus a most powerful $\powerGnn$ with the recursion in~\Eqref{eq:WL} will map them to distinct collections of vertex embeddings.  Thus, the  graph embeddings $\bh_{G_{l}}$, obtained by applying an injective read-out function to $\big\{\big\{\bh^{(1)}_{l, v} \big\}\big\}_{v \in V_{l}}$ will be distinct: $\bh_{(G_{1})_{\pi, \pi}} \ne \bh_{(G_{2})_{\pi', \pi'}}$, as desired.


As no representation of $G_2$ can match any representation of $G_1$, we can find a function $g(\cdot)$ that, when composed with $\powerGnn$, ensures that the sum in \Eqref{eq:jointinv} gives different values for $G_1$ and $G_2$ by Lemma~5 of \citet{xu2018how} (or Theorem~2 of \citet{Zaheer2017}). Since we can always redefine $\powerGnn{}' = g \circ \powerGnn$ and $\powerGnn{}'$ is still a WL-GNN, we conclude our proof.
\end{proof}
\section*{Proposition~\ref{p:piSGD}}
The following proposition regarding the convergence of $\pi$-SGD was stated in the paper:
\pisgd*

Here we list the relevant conditions.  \citet{murphy2018janossy} point out that $\pi$-SGD can be characterized by the work of~\citet{younes1999convergence, Yuille2004}  
  and is a a familiar application of stochastic approximation algorithms already used in training neural networks.

In particular, the following assumptions are made:

\begin{enumerate}
	\item There exists a constant $M > 0$ such that for all $\mW$, $- \mG_t^T \mW \leq M \Vert \mW - \mW^\star \Vert_2^2$,
	where $\mG_t$ is the true gradient for the full batch over all permutations and $\mW^\star$ is an optimum.
	
	\item there exists a constant $\delta > 0$ such that for all $\mW$, $\expected_t [ \Vert \rmZ_t\Vert_2^2] \leq \delta^2 (1 +\Vert \mW_{t} - \mW^{\star}_{t} \Vert_2^2)$, where $\rmZ_t$ is the random gradient of the loss w.r.t. weights at step $t$ and the expectation is taken with respect to all the data prior to step $t$.
\end{enumerate}

If these assumptions are satisfied, then $\pi$-SGD (as with SGD) converges to a fixed point with probability one.

\section*{Proof of Proposition~\ref{prop:kPerms}}
We restate the proposition for completeness.
\begin{proposition*}
The RP in \Eqref{eq:karyAH} requires summing over all $k$-node induced subgraphs of $G$, thus saving computation when $k < |V|$, reducing the number of terms in the sum from $|V|!$ to $\frac{|V|!}{(|V| - k)!}$.
\end{proposition*}
\begin{proof}
$k$-ary RP needs to iterate over the $k$-node induced subgraphs of $G$ ($\binom{|V|}{k}$ subgraphs), but for each subgraph there are $k!$ different ways to order its nodes, resulting in $\frac{|V|!}{(|V| - k)!}$ evaluations of $\harrow{f}$. 
\end{proof}

\section*{Proof of Proposition~\ref{thm:k_implies_km}}
We restate the proposition for completeness.
\begin{proposition*}
$\dbar{f}^{(k)}$ becomes strictly more expressive as $k$ increases.
That is, for any $k\in\mathbb{N}$, define $\mathcal{F}_{k}$ as the set of all permutation-invariant graph functions that can be represented by RP with $k$-ary dependencies. 
Then, $\mathcal{F}_{k-1}$ is a \emph{proper} subset of $\mathcal{F}_k$. 
Thus, RP with $k$-ary
dependencies can express any RP function with $(k-1)$-ary
dependencies, but the converse does not hold.
\end{proposition*}
\begin{proof}
($\mathcal{F}_{k-1}\subset\mathcal{F}_{k}$): %
Consider an arbitrary element 
$\dbar{f}^{(k-1)} \in \mathcal{F}_{k-1}$, and write $\harrow{f}(\tA[1\!\!:\!\!(k\!-\!1),1\!\!:\!\!(k\!-\!1),:], \Xv[1\!\!:\!\!(k\!-\!1),:]; \mW)$ for its 
associated permutation-sensitive RP function.
Also consider $\dbar{f}^{(k)} \in \mathcal{F}_{k}$ and let $\harrow{f}'$ be its associated permutation-sensitive RP function.
For any tensor $\tA$ and attribute matrix $\Xv$, 
we can define $\harrow{f}'(\tA[1\!\!:\!\!k,1\!\!:\!\!k,:], \Xv[1\!\!:\!\!k,:]; \mW) = \harrow{f}(\tA[1\!\!:\!\!(k\!-\!1),1\!\!:\!\!(k\!-\!1),:], \Xv[1\!\!:\!\!(k\!-\!1),:]; \mW)$. Thus, $\dbar{f}^{(k-1)} \in \mathcal{F}_{k}$ and because $\dbar{f}^{(k-1)}$ is arbitrary, we conclude $\mathcal{F}_{k-1}\subset\mathcal{F}_{k}$.

~\\
($\mathcal{F}_{k}\not\subset\mathcal{F}_{k-1}$): %
The case where $k=1$ is trivial, so assume $k > 1$.  
We will demonstrate  $\exists \dbar{f}^{(k)}\in\mathcal{F}_{k}$ such that $\dbar{f}^{(k)} \not \in \mathcal{F}_{k-1}$.  Let $\dbar{f}^{(k)}$ and $\dbar{f}^{(k-1)}$ be associated with $\harrow{f}^{(k)}$ and $\harrow{f}^{(k-1)}$, respectively.

\paragraph{Task.}

Consider the task of representing the class of circle graphs with skip links shown in \Figref{f:circlegraphs}.  Let $G_{k} \in \skipGraph(\skipNumVert_{k}, k)$ and $G_{k+1} \in \skipGraph(\skipNumVert_{k}, k+1)$ where $\skipNumVert_{k}$ is any prime number satisfying $\skipNumVert_{k} > 2(k-1)(k+1)$.  That is, $G_{k}$ and $G_{k+1}$ are circulant skip length graphs with the same number of vertices and skip lengths of $k$ and $k+1$, respectively.  Note that $\skipNumVert_{k} > k + 1$ is prime and thus it is co-prime with both $k$ and $k+1$; further, $\skipNumVert_{k} - 1> k + 1$ so the conditions for creating the CSL graph in Definition~\ref{def:circleGraph} are indeed satisfied.  To complete the proof, we need to show that (1) there is a $\dbar{f}^{(k)}$ capable of distinguishing $G_{k}$ from $G_{k+1}$ but (2) no such $\dbar{f}^{(k-1)}$ exists.

Denote $G_{\skipLen} = (\mA_{\skipLen}, \Xv)$, $\skipLen \in \{k, k+1 \}$, where $\Xv = c \boldsymbol{1}$ for some $c \in \mathbb{R}$ for both graphs, as there are no vertex features, and where  $\mA_{\skipLen}$ represents an adjacency matrix for $G_{
\skipLen}$ (there are no edge features).

\paragraph{(1) A $k$-ary $\dbar{f}^{(k)}$ that can distinguish between $G_{k}$ and $G_{k+1}$.}

We will define $\harrow{f}^{(k)}$ in terms of a composition with a canonical orientation. 
In particular, we only allow the orientation of $\mA$ (and $\Xv$) that arises from  first generating an edgelist by the scheme described in Definition~\ref{def:circleGraph} and then constructing the adjacency matrix from it in the usual way.   
For either graph, we define
$\harrow{f}^{(k)}((\mA_{\skipLen})_{\pi, \pi}[1:k, 1:k, :], \Xv_{\pi}[1:k, :]) = 0$ for all permutations that do not yield this `canonical' adjacency matrix.   

Under this orientation,  the $k \times k$ submatrix $\mA_{k}[1:k, 1:k]$ of $G_k$ will have more nonzero elements than that of $G_{k+1}$, $\mA_{k+1}[1:k, 1:k]$.  The relevant induced subgraph of size $k$ in $G_k$ will include a pair of vertices that are $k$ `hops' away which will thus be connected by an edge, whereas in $G_{k+1}$, the skip length is too long so its induced subgraphs of size $k$ will have fewer edges.  Therefore, it suffices to let $\harrow{f}^{(k)}$ count the number of nonzero elements in the (properly oriented) submatrices presented to it.

\paragraph{(2) No $(k-1)$-ary $\dbar{f}^{(k-1)}$ can distinguish between $G_{k}$ and $G_{k+1}$.}

We will show that the induced subgraphs of size $k-1$ are ``the same'' in both $G_{k}$ and $G_{k+1}$, which will imply that no satisfactory $\harrow{f}^{(k-1)}$ can be constructed.  In particular, if we denote by $\sgListKay$ the multiset of induced subgraphs of size $k-1$ in $G_{k}$ and by $\sgListKpo$ the multiset of induced subgraphs of size $k-1$ in $G_{k+1}$, it can be shown that $\sgListKay$ and $\sgListKpo$ are equivalent in the following sense.  There exists a bijection $\phi$ between these finite multisets such that for every $H \in \sgListKay$, $\phi(H) \in \sgListKpo$ is isomorphic to $H$.  For example, the multisets  (with arbitrarily labeled vertices) $\big\{\big\{$ \inlinePath{1}{2}{3}, \inlinePath{2}{3}{4}, \inlineTriangle{5}{6}{7} $\big\}\big\}$ and 
$\big\{\{$ \inlinePath{1}{3}{5}, \inlinePath{2}{5}{7}, \inlineTriangle{5}{7}{9} $\big\}\big\}$ have an isomorphism-preserving bijection between them and will thus be considered equal.

In the interest of brevity, what follows is a sketch.  We first observe that we only need to consider the multisets of induced subgraphs that include vertex $0 \in V = \{0, 1, \ldots, \skipNumVert_{k}-1 \}$ due to the vertex transitivity of the CSL graphs.  Next, we observe that we only need to consider the multisets of `maximally connected' induced subgraphs of size $k-1$.  By `maximally connected', we mean an induced subgraph of size $k-1$ such that no more edges from $G_{\skipLen}$, $\skipLen \in \{k , k+1 \}$, can be added without adding a $k^{\text{th}}$ vertex to the induced subgraph. 
Indeed, once we show that a bijection exists between maximally connected induced subgraphs of size $k-1$, it follows that such a bijection exists for \emph{any} connected induced subgraphs of size $k-1$ since these can be formed by deleting any edge that does not render the induced subgraph disconnected.  Then, viewing disconnected graphs as the disjoint union of connected components, a similar argument to the one applied for connected induced subgraphs can be used to complete the argument for any possible induced subgraph of size $k-1$.

We can construct all such  maximally connected subgraphs including $0 \in V$ in both $G_{\skipLen}$ for $\skipLen \in \{k, k+1 \}$ by forming recursive sequences on the integers $\{0, 1, \ldots, \skipNumVert_{k} - 1\}$ with addition mod $\skipNumVert_{k}$ (see for instance Definition~\ref{def:circleGraph}); the key difference in these sequences is whether $\skipLen = k$ or $\skipLen = k+1$ can be added to or subtracted from the previous value in the sequence.  We will call the former a $k$-sequence and the latter a $(k+1)$-sequence.  In either case, distinct sequences  may result in the same induced subgraphs but we can simply take one representative from each equivalence class.  

Importantly, these sequences can be constructed in a way that abstracts from either underlying graph $G_{k}$ or $G_{k+1}$.  Due to our choice of $\skipNumVert_{k}$, the recursive sequences never `wrap around' the graph and can be informally thought of as a recursive sequence on the integers of a bounded interval with 0 in the middle rather than a circle with skip links.  In particular, we can define 
recursive sequences on the set of integers  $\{-(k+1)(k-1), \ldots, (k+1)(k-1) \}$ with regular addition to construct the same induced subgraphs (-1 corresponds to vertex $(\skipNumVert_{k}-1)\in V$ and so on, in either case).  Then it becomes clear that there is an isomorphism-preserving bijection between the induced subgraphs formed by recursive $(k+1)$-sequences and $k$-sequences on this bounded interval of integers; any sequence defined in terms of adding or subtracting $k+1$ can be replaced by one that adds or subtracts $k$ (and vice versa), which completes the proof.

\end{proof}

%
%
\section{Further Details of the Experiments}
\label{appendix:experiments}
\subsection{\AH Pooling and Graph Structure Representation on CSL Graphs}
Our GIN architecture uses five layers of recursion, where every $\mathrm{MLP}^{(l)}$ has two hidden layers with 16 neurons in the hidden layers.  The graph embedding is mapped to the output through a final linear layer $\mathrm{softmax}(\bh_{G}^{T}\bW)$.   $\epsilon^{(l)}$ is treated as a learnable parameter.  With standard GIN, since the vertex attributes are not one-hot encoded (they are constants), we first apply an MLP embedding before computing the first update recursion (as in~\citet{xu2018how}).  Since RP-GIN utilizes one-hot IDs, we do not need an MLP embedding in the first update.  For these experiments, we assign one-hot encoding of $i \ \mathrm{mod} \ 10$ for $i \in \{1,2 \ldots, |V|=41 \}$ -- rather than completely unique IDs -- which facilitates learning.  We train with $\pi$-SGD, applying one random permutation of each adjacency matrix at each epoch.  For inference, we average the score over five random permutations of each graph, as in Remark~\ref{rmk:inference}.   \Figref{fig:visEmbeddings} shows the stronger performance of RP-GIN on this task. 

Both models are trained for 1000 epochs using ADAM~\citep{KingmaAdam} for optimization.

For the cross-validation, we use five random initializations at each fold.  The folds are such that the classes are balanced in both training and validation.

Models are trained on CPUs but on machines with multiple CPUs; PyTorch inherently multithreads the execution.

\begin{figure}[t!!!]
	\begin{center}
		\includegraphics[height=2.7cm, width=3cm]{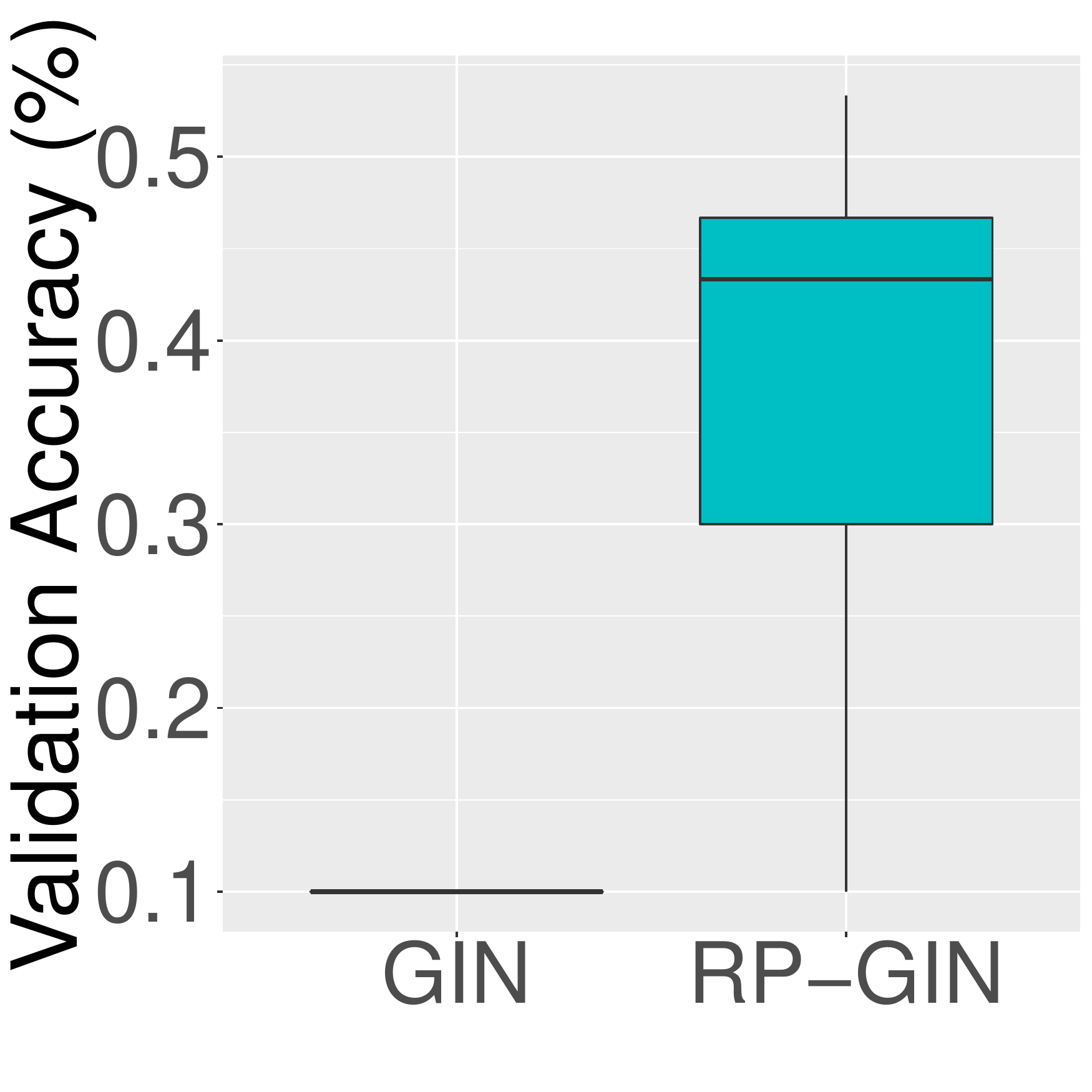}
	\end{center}
	\caption{RP-GNN is more powerful than WL-GNN in a challenging 10-class classification task. 
		\label{fig:visEmbeddings}}
\end{figure}
\subsection{Predicting Molecular Properties}
Here we provide additional details on the molecular experiments. 
 (1) For the models based on Graph Convolution~\citep{Duvenaud2015, altae2017low}, we extend the architecture provided from DeepChem and the MoleculeNet project.  Following them, the learning rate was set to 0.003, we trained with mini-batches of size 96, and used the Adagrad optimizer\citep{duchi2011adaptive}.  Models were trained for 100 epochs.  Training was performed on 48 CPUs using the inherent multithreading of DeepChem.
\begin{table}
	\caption{Datasets used in our experiments. }
	\label{tab:datasets}
	\scalebox{0.9}{
		\begin{tabular}{lrr}
			Data Set & Number of Compounds & Number of Tasks \tabularnewline
			\hline
			HIV & 41,127 & 1 \tabularnewline
			MUV & 93,087 &  17 \tabularnewline
			Tox 21 & 6,284  & 12 \tabularnewline
		\end{tabular}
	}
\end{table}

Note that we re-trained DeepChem models using this fewer number of epochs to make results comparable. That being said, many models reached optimal performance before the last epoch; we use the model with best validation-set performance for test-set prediction.

(2) For the so-called RNN and CNN models, all MLPs have one hidden layer with 100 neurons.  We used the Adam optimizer~\citep{KingmaAdam}, again training all models with mini-batches of size 96 and 50 epochs.  We performed a hyperparameter line search over the learning
rate, with values in $\{0.003, 0.001, 0.01, 0.03, 0.1, 0.3\}$.  Training was performed on GeForce GTX 1080 Ti GPUs.
To model the RNN, we use an LSTM with 100 neurons and use the long term memory as output.

\paragraph{RP-Duvenaud}
When we train RP-Duvenaud, we follow any training particulars as in the DeepChem implementation.  For instance, DeepChem's implementation computes a weighted loss which penalizes misclassification differently depending on the task, and they compute an overall performance metric by taking the mean of the AUC across all tasks (see Table~\ref{tab:datasets}).  One difference is that the DeepChem recommends either metrics PRC-AUC or ROC-AUC and splits ``random'' or ``scaffold'' depending on the dataset under consideration.  Since ROC-AUC and random splits were the most commonly used among the three datasets we chose, we decided -- before training any models -- to use random splits and ROC-AUC for every dataset for simplicity.  We also note that the authors of MoleculeNet report ROC-AUC scores on all three datasets.  Regarding the sizes of the train/validation/test splits, we used the default values provided by DeepChem.

We implement the model that assigns unique IDs to atoms by first finding the molecule with the most atoms across training, validation, and test sets, and then appending a feature vector of that size to the endowed vertex attributes.  That is, if the largest molecule has $A$ atoms, we concatenate a vector of length $A$ of one-hot IDs to the existing vertex attributes (for every vertex in each molecule). 

\paragraph{CNNs and RNNs}
We explore $k=20$-ary \AHP with $\harrow{f}$ as a CNN, learned with $\pi$-SGD.  At each forward step, we run a DFS from a different randomly-selected vertex to obtain a $20 \times 20 \times 14$ subtensor of $\tA$ (there are 14 edge features), which we feed through two iterations of $\text{conv} \rightarrow \text{ReLU} \rightarrow \text{MaxPool}$ to obtain a representation $\vh_{\tA}$ of $\tA$.  The corresponding vertex attributes are fed through an MLP and concatenated with $\vh_{\tA}$ to obtain a representation $\vh_{G}$ of the graph which in turn is fed through an MLP to obtain the predicted class (see also~\Eqref{eq:RPCNN}).  Zero padding was used to account for the variable-sized molecules.  Twenty initial vertices for the DFS (i.e. random permutations) were sampled at inference time. Table~\ref{tab:molaucs} shows that the CNN $\harrow{f}$ underperforms in all tasks. 

We also consider RP with an RNN as $\harrow{f}$ learned with $\pi$-SGD, starting with a DFS to yield a $|V|$ $\times$ $|V|$ $\times$ 14 subtensor.  For $\harrow{f}$, we treat the edge features of a given vertex as a sequence: for vertex $v$, we apply an LSTM to the sequence $(\tA_{v, 1, \cdot}, \tA_{v, 2, \cdot}, \ldots, \tA_{v, |V|, \cdot})$ and extract the long-term state.  We also take the vertex attributes and pass them through an MLP.  The long term state and output of the MLP are concatenated, ultimately forming a representation for every vertex (and its neighborhood) which we view as a second sequence.  We apply a second LSTM and again extract the long term state, which can be denoted $\vh_{G}$, the embedding of the graph.  Last, $\vh_{G}$ is forwarded through an MLP yielding a class prediction.  Twenty starting vertices (i.e. permutations) were sampled at inference time.  Variability was quantified with 5 random train/val/test splits for both neural network based models. Interestingly, Table~\ref{tab:molaucs} shows that the RP-RNN approach performs reasonably well in the Tox21 dataset, while underperforming in other datasets. Future work is needed to determine those tasks for which the RNN and CNN approaches are better suited.

\paragraph{$\pi$-SGD}

To train with $\pi$-SGD, we sample a different random permutation of the graph at each forward pass.  In the case of RP-Duvenaud, this involves assigning permutation-dependent unique IDs at each forward step (as in~\Eqref{eq:RPGNN}).  In our implementation, we achieve this  by building a new DeepChem object for the molecule at each forward pass.  This operation is expensive but we did not consider refined code optimizations for this work.  
In general, with properly optimized code, sampling permutations need not be as expensive and allows for a tractable and theoretically justified procedure.
Looking ahead to the test data in order to find the largest molecule in test and validation corresponds to using domain knowledge and the modeling choice that the resulting model will only work on molecules with at most $A$ atoms.  It is not hard to construct a similar model that does not rely on this look-ahead mechanism, such as assigning a one-hot encoding of $i~\mathrm{ mod }~ A_{\text{observed}}$ where $i \in \{1, 2, \ldots, |V| \}$ and $A_{\text{observed}}$ is the largest molecule observed in the model building phase.

\paragraph{Molecule dataset details}
Details on the molecular datasets are shown in Table~\ref{tab:datasets}. The observant reader may notice that we report a different number of Tox21 molecules than in~\citet{wu2018moleculenet}.  This resulted from simultaneously finding (1) that the validation and testing Python objects for Tox21 were empty when we first loaded them in the early stage of development and (2) comments in the DeepChem source code that led us to believe that this was expected behavior.  We thus split up the `training' dataset rather than using the provided splits.  This treatment was the same for all models, making the comparison fair.

The number of molecules in each dataset with greater than $k = 10, 20, 30, 40, 50$ molecules are 98.18\%, 63.71\%, 22.30\%, 7.71\%, 3.59\% for HIV; 99.93\%, 75.33\%, 12.30\%, 0.03\%, 0.00\% for MUV; and 78.07\%, 33.63\%, 10.39\%, 3.90\%, 1.97\% for Tox21.


%

\begin{figure*}[t!!!]

		\centering
		\includegraphics[scale=0.55]{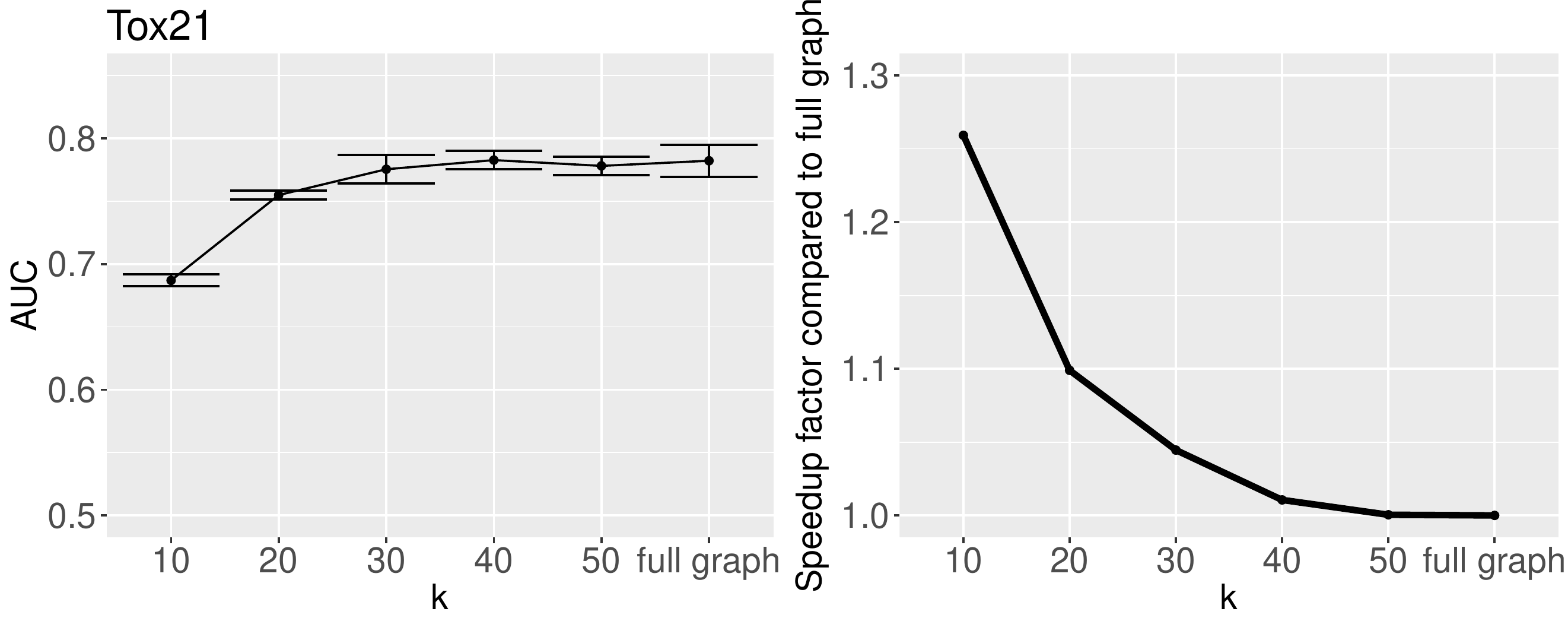}
			\caption{Training time and model performance of $k$-ary models for the Tox21 task.  Test-set AUC was computed for five different random splits of train/validation/test: we show the mean $\pm$ one standard deviation.  We also show the speedup factor for training: time to train on the full graph divided by time for $k$-ary model.  Training was performed on 48 CPUs, making use of PyTorch's inherent multithreading.}
			
			\label{fig:toxfig}
\end{figure*}
\begin{figure*}[t!!!]

		\centering
		\includegraphics[scale=0.5]{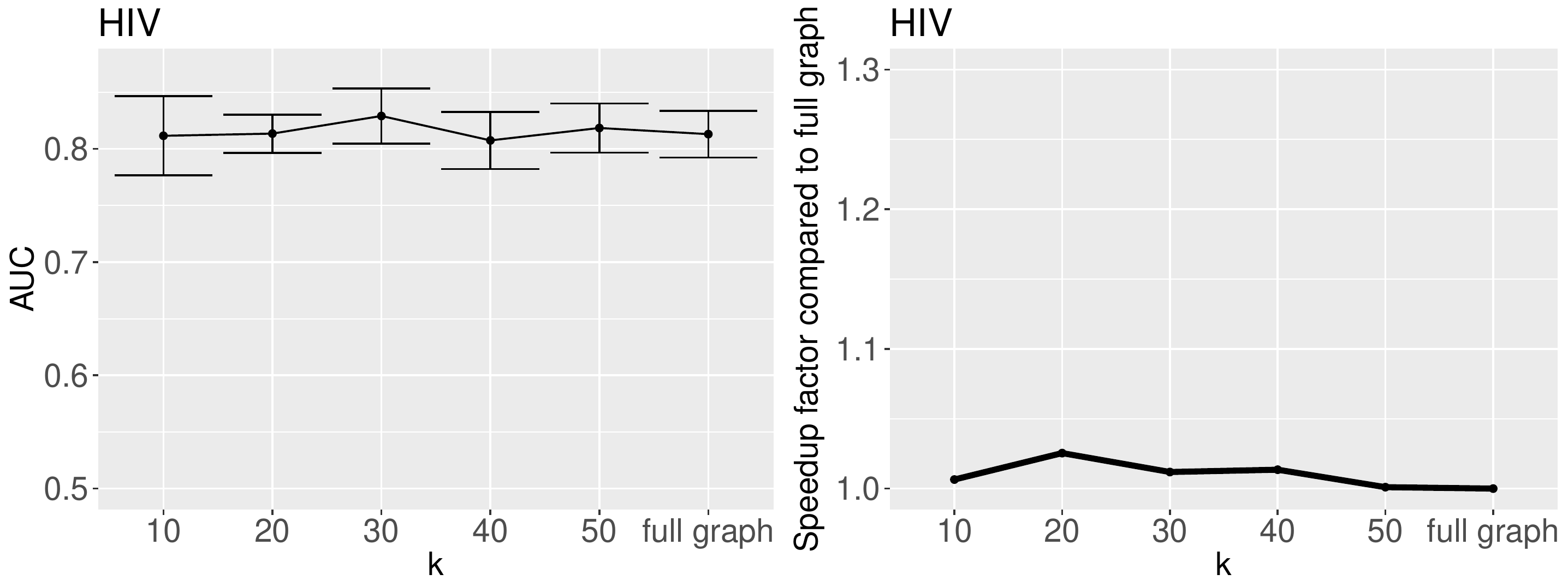}
		\caption{Training time and model performance of $k$-ary models for the HIV task. }
			\label{fig:hivfig}
\end{figure*}

\begin{figure*}[b]

	\centering
	\includegraphics[scale=0.5]{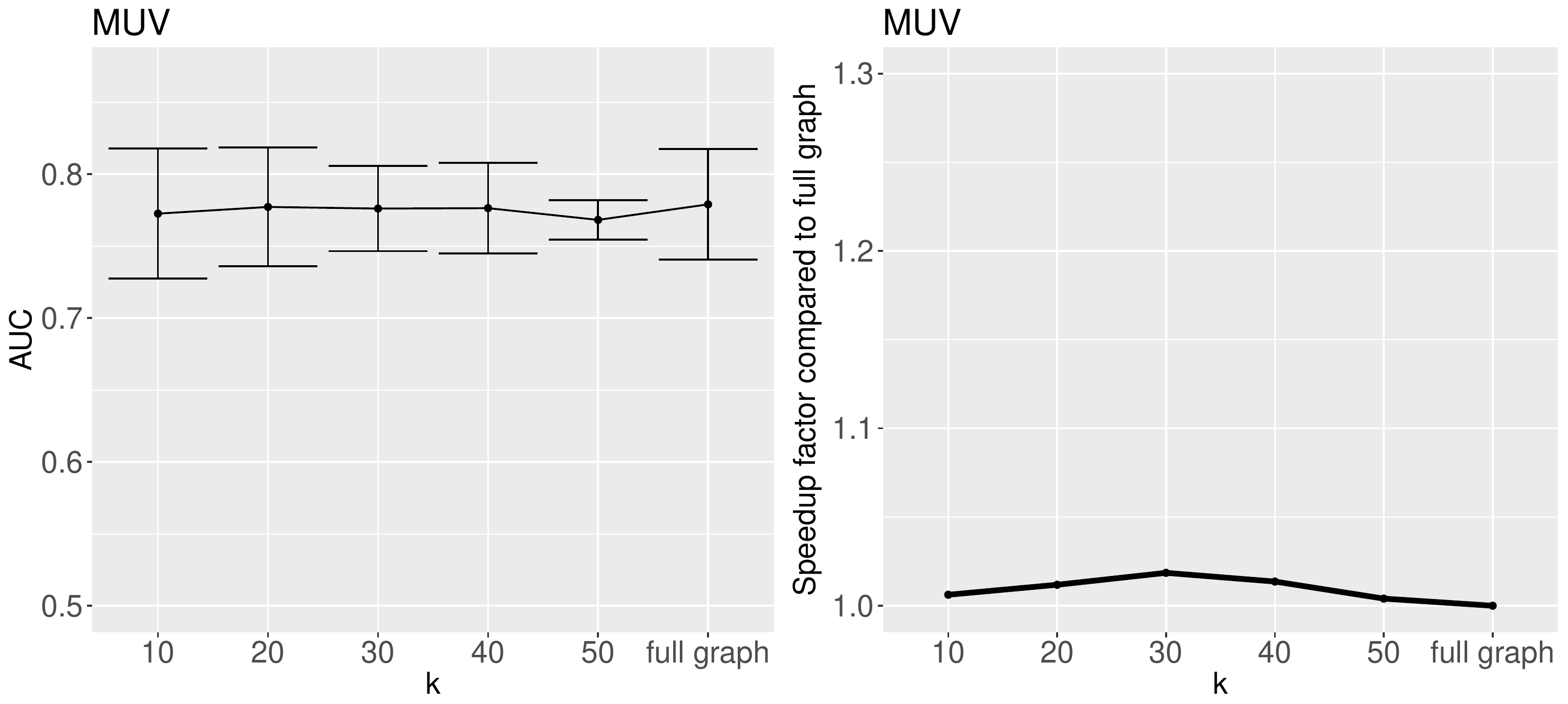}
	\caption{Training time and model performance of $k$-ary models for the MUV task. }
		\label{fig:muvfig}
\end{figure*}

\end{document}